\documentclass[journal]{IEEEtai}

\usepackage[colorlinks,urlcolor=blue,linkcolor=blue,citecolor=blue]{hyperref}

\usepackage{color,array}

\usepackage{graphicx}
\usepackage{makecell}
\usepackage{array}
\usepackage{balance}
\usepackage{amsmath}
\usepackage{graphicx}
\graphicspath{{Figures/}}
\usepackage{pdfpages}
\usepackage{cite}
\usepackage{lipsum}
\usepackage{subfig} 
\usepackage[T1]{fontenc} 

\usepackage{url}

\usepackage{wrapfig}
\usepackage{lipsum}

\usepackage{multirow}
\usepackage{multicol}
\usepackage{longtable}
\usepackage{tabularx}
\usepackage{booktabs}
\usepackage{mathtools}

\usepackage{gensymb} 

\usepackage[font=scriptsize,labelfont=bf,tableposition=top]{caption}
\usepackage{amsmath}
\usepackage{amssymb}  
\usepackage{amsfonts}
\usepackage{float}       
\usepackage{algorithm}   
\usepackage{algpseudocode} 
\usepackage{setspace}
\newcolumntype{?}{!{\vrule width 1pt}}
\usepackage{hyperref}
\usepackage{adjustbox}

\begin{document}

\title{NeuroDDAF: Neural Dynamic Diffusion-Advection Fields with Evidential Fusion for Air Quality Forecasting} 

\author{
\thanks{Manuscript received 10-Feb-2026; revised Month-Day-2026; accepted Month Day 2026.}  
}

 \author{ Prasanjit~Dey,
     Soumyabrata~Dev, 
     Angela~Meyer, and
      Bianca~Schoen-Phelan 
 }

\markboth{IEEE Transactions on Artificial Intelligence ,~Vol.~XX, No.~XX, XX~2026}%
{IEEE Transactions on Artificial Intelligence ,~Vol.~XX, No.~XX, XX~2026}

\maketitle

\begin{abstract}
Accurate air quality forecasting is crucial for protecting public health and guiding environmental policy, yet it remains challenging due to nonlinear spatiotemporal dynamics, wind-driven transport, and distribution shifts across regions. Physics-based models are interpretable but computationally expensive and often rely on restrictive assumptions, whereas purely data-driven models can be accurate but may lack robustness and calibrated uncertainty. To address these limitations, we propose Neural Dynamic Diffusion-Advection Fields (NeuroDDAF), a physics-informed forecasting framework that unifies neural representation learning with open-system transport modeling. NeuroDDAF integrates (i) a GRU--Graph Attention encoder to capture temporal dynamics and wind-aware spatial interactions, (ii) a Fourier-domain diffusion-advection module with learnable residuals, (iii) a wind-modulated latent Neural ODE to model continuous-time evolution under time-varying connectivity, and (iv) an evidential fusion mechanism that adaptively combines physics-guided and neural forecasts while quantifying uncertainty. Experiments on four urban datasets (Beijing, Shenzhen, Tianjin, and Ancona) across 1-3 day horizons show that NeuroDDAF consistently outperforms strong baselines, including AirPhyNet, achieving up to 9.7\% reduction in RMSE and 9.4\% reduction in MAE on long-term forecasts. On the Beijing dataset, NeuroDDAF attains an RMSE of 41.63~$\mu$g/m$^3$ for 1-day prediction and 48.88~$\mu$g/m$^3$ for 3-day prediction, representing the best performance among all compared methods. In addition, NeuroDDAF improves cross-city generalization and yields well-calibrated uncertainty estimates, as confirmed by ensemble variance analysis and case studies under varying wind conditions.
\end{abstract}

\begin{IEEEImpStatement}
Air pollution forecasting underpins health advisories and mitigation planning, but real-world deployments must handle rapidly changing meteorology, sparse sensing, and shifts across locations. This paper presents NeuroDDAF, a physics-informed forecasting framework that combines diffusion-advection transport structure with learned spatiotemporal representations and calibrated uncertainty. By explicitly modeling wind-driven transport and adaptively fusing physics-guided and neural predictions through an evidential gate, NeuroDDAF improves accuracy and robustness across multiple cities and lead times while providing prediction intervals for risk-aware decisions. These capabilities can support more dependable early warnings and policy-relevant forecasting, and more broadly illustrate how integrating physical priors with uncertainty-aware learning can improve the trustworthiness and transferability of AI forecasting systems in safety- and decision-critical settings.
\end{IEEEImpStatement}

\begin{IEEEkeywords}
Air quality forecasting, physics-informed machine learning, diffusion-advection, spatiotemporal graph neural networks, uncertainty quantification.
\end{IEEEkeywords}

\section{Introduction} \label{sec:introduction}

\IEEEPARstart{A}{ir} pollution remains one of the most pressing environmental challenges of our time. It poses a significant threat to public health, ecosystems, and the global climate~\cite{azimi2024unveiling}. Recent WHO reports indicate that 9 out of 10 people worldwide breathe polluted air, contributing to 7 million premature deaths annually~\cite{kan2023air}. Accurate air quality forecasting plays a critical role in enabling timely mitigation strategies, issuing public health advisories, and guiding environmental policy. However, the task is highly challenging due to the complex, nonlinear, and spatiotemporally dynamic nature of pollutant dispersion, which is influenced by meteorological conditions, urban topologies, and emission sources.

Traditional approaches to air quality prediction mainly rely on physics-based models and data-driven methods~\cite{liao2020deep}. However, each of these approaches has its inherent limitations. Physics-based models are primarily inspired by atmospheric science and domain-specific knowledge, aiming to formulate mathematical equations that simulate the processes governing air pollution. For example, Chemical Transport Models (CTMs), rely on solving complex differential equations to simulate atmospheric processes like diffusion and advection~\cite{li2023physics}. While physics-based models offer the advantage of producing interpretable results grounded in well-established physical laws and atmospheric principles, they are not without significant limitations. One of the primary challenges is that these models often rely on rigid assumptions. For example, they may treat the atmosphere as a closed or near-equilibrium system. This limits their ability to capture the dynamic and non-linear nature of real-world conditions. For instance, they frequently struggle to account for sudden changes in pollutant emissions, unexpected meteorological disturbances, or localized phenomena that deviate from average regional trends~\cite{bauer2015quiet}. Moreover, most physics-based approaches require extensive parameter calibration, often relying on historical data and expert tuning to ensure accurate simulations. This calibration process can be both time-consuming and inflexible, making it difficult to adapt these models to rapidly changing environments or to incorporate complex, unpredictable human behaviors. Consequently, the real-time applicability and generalizability of such models remain limited in urban or industrial areas.

On the other hand, purely data-driven deep learning models rely on a few years of historical pollution and meteorological data to learn complex relationships without incorporating knowledge of the underlying physical processes. Among recent approaches, deep learning models particularly Graph Neural Networks (GNNs)~\cite{du2025graph} and Transformer-based architectures have shown strong potential due to their expressive modeling capacity. For instance, Spatiotemporal Graph Neural Networks (STGNNs) integrate Recurrent Neural Networks (RNNs) with GNNs to jointly capture temporal sequences and spatial dependencies, leading to improved prediction accuracy~\cite{ji2023spatio}. Meanwhile, attention-based models have emerged as promising alternatives for long-term air quality forecasting~\cite{wang2022air, liang2023airformer}, effectively modeling complex spatiotemporal dependencies. Despite these advances, most of these models remain agnostic to physical laws, resulting in limited interpretability, reduced robustness under distribution shifts, and unreliable performance in sparse or extrapolation scenarios.

Recent studies have tried to close the gap between traditional physics models and deep learning by adding physical knowledge into neural networks. One well-known study is AirPhyNet~\cite{hettige2024airphynet}, which uses diffusion-advection equations inside a neural ODE framework to model how pollutants move through the air. This approach led to strong results and set a new standard for air quality prediction. Although AirPhyNet achieves strong results, it is limited by several key assumptions. First, it treats the system as closed, meaning that pollutants are assumed to remain within a fixed domain. This assumption does not hold in real-world environments, where wind-driven advection, external emissions, and natural deposition all contribute to pollutant exchange across boundaries. Second, AirPhyNet represents physical processes in latent space, making them difficult to interpret and weakly tied to actual physics. Third, it relies on static graph structures and fixed fusion mechanisms, which fail to account for dynamic meteorological changes or uncertainty in predictions. So while AirPhyNet performs well on some datasets, its design makes it harder to apply in new environments or under changing conditions.

To address these limitations, we propose Neural Dynamic Diffusion-Advection Fields (NeuroDDAF), a forecasting framework that tightly integrates data-driven learning with physical modeling in an open-system setting. NeuroDDAF is explicitly designed to capture the real-world dynamics of air pollution transport across both space and time, while remaining flexible to heterogeneous regions and data regimes. The framework is built around four central innovations:

\begin{itemize}
\item \textbf{Spatiotemporal Encoding with GRU-GAT:} A hybrid encoder couples Gated Recurrent Units (GRUs) for temporal sequence modeling with a wind-aware Graph Attention Network (GAT) for spatial interaction learning. This enables joint representation of local temporal dependencies and dynamic spatial effects driven by meteorological conditions.
\item \textbf{Fourier-Domain PDE Solver:} To incorporate transport physics, we design a Fourier-based Partial Differential Equations (PDE) solver that augments classical advection-diffusion dynamics with learnable neural residuals. This solver efficiently captures both smooth diffusion and anisotropic advection in the frequency domain, preserving physical interpretability.
\item \textbf{Latent Neural ODE with Dynamic Wind Coupling:} We model pollutant trajectories in latent space using a neural ODE whose dynamics are modulated by real-time wind fields and evolving graph Laplacians. Unlike static graph structures, this formulation accounts for time-varying connectivity induced by meteorological variability.
\item \textbf{Evidential Fusion and Uncertainty Quantification:} Finally, we introduce an evidential gating mechanism that adaptively fuses physics-based PDE predictions with neural forecasts according to confidence scores. This is paired with evidential uncertainty estimation, providing calibrated prediction intervals and robustness under abrupt pollution events.
\end{itemize}

We validate NeuroDDAF on four large-scale urban air quality datasets. The model consistently outperforms AirPhyNet and other strong baselines in terms of Mean Absolute Error (MAE) and Root Mean Squared Error (RMSE), while demonstrating improved generalization and uncertainty calibration across diverse environments.

\section{Related Works}

Air quality prediction has been extensively studied across environmental science, atmospheric modeling, and machine learning communities, given its profound implications for public health and urban sustainability. Existing approaches can be broadly categorized into three classes: (i) physics-based models, (ii) purely data-driven deep learning models, and (iii) hybrid physics-guided deep learning frameworks, each addressing distinct challenges inherent in air pollution modeling.
\subsection{Physics-Based Models}
Traditional physics-based models primarily rely on solving PDEs such as the advection-diffusion equation to simulate pollutant transport and transformation under meteorological influences. These models encode established atmospheric principles and offer interpretability grounded in domain knowledge~\cite{daly2007air,li2023physics}. However, they suffer from high computational costs, rigid closed-system assumptions, and sensitivity to parameter calibration errors, limiting real-time applicability in dynamic urban environments~\cite{bauer2015quiet,necati1994finite}. 

\subsection{Data-Driven Approaches}

The advent of data-driven methods for air quality prediction, leveraging historical pollutant and meteorological datasets to model nonlinear spatiotemporal dependencies without explicit physics prior~\cite{han2021joint, han2023kill}. Traditional shallow machine learning approaches~\cite{sanchez2013nonlinear, su2023novel} often struggle to capture the intricate nonlinear dependencies inherent in air quality dynamics. In contrast, recent advances in deep learning techniques~\cite{luo2019accuair, wang2022air} have demonstrated notable improvements by effectively modeling complex spatiotemporal correlations across monitoring networks. For example, Transformer-based models like AirFormer~\cite{liang2023airformer} introduced multi-head attention mechanisms to capture long-range correlations across thousands of monitoring stations at nationwide scales while incorporating uncertainty modeling through stochastic latent variables. Despite these gains, such deep models typically demand large volumes of training data to achieve high predictive accuracy and remain largely agnostic to physical principles. This absence of physics integration not only limits their ability to generalize under distribution shifts or unseen environmental conditions but also hampers their interpretability in real-world atmospheric contexts.

\subsection{Physics-Guided Deep Learning}

Recently, several studies have introduced physics-guided deep learning approaches that integrate explicit physical knowledge into neural architectures, thereby enhancing robustness, interpretability, and data efficiency~\cite{raissi2017physics,peng2022physics}. While some approaches integrate physical constraints directly into the loss function, others adopt hybrid network architectures that combine data-driven learning with physics-based modeling. For instance, Jia~\textit{et~al.}~\cite{jia2019physics} proposed a physics-informed loss function to enhance the accuracy of lake temperature prediction. Similarly, Wang~\textit{et~al.}~\cite{wang2020towards} developed a composite framework for turbulent flow forecasting that leverages fundamental physical properties of the process. In the context of traffic modeling, Ji~\textit{et~al.}~\cite{ji2020interpretable} and Wang~\textit{et~al.}~\cite{wang2022traffic} introduced interpretable models for grid-based traffic flow prediction using potential energy field representations, while Ji~\textit{et~al.}~\cite{ji2022stden} extended this idea by presenting a physics-guided neural network tailored for road network traffic flow prediction. Furthermore, Mohammadshirazi~\textit{et~al.}~\cite{mohammadshirazi2023novel} introduced a hybrid framework that integrates physics-based state-space models with machine learning techniques to improve indoor air quality prediction. More recently, physics-guided deep learning has been applied to outdoor air quality prediction. Hettige~\textit{et~al.}~\cite{hettige2024airphynet} introduced AirPhyNet, embedding diffusion-advection processes into a graph-based differential equation network for improved long-term and sparse-data forecasting. Liang~\textit{et~al.}~\cite{liang2023airformer} proposed AirFormer, a Transformer-based nationwide predictor that incorporates domain knowledge via specialized spatial-temporal attention mechanisms and a stochastic stage to model uncertainty. Tian~\textit{et~al.}~\cite{tian2024air} developed Air-DualODE, which combines a Boundary-Aware Diffusion-Advection Equation for open-system dynamics with a complementary data-driven branch, aligning and fusing both to achieve state-of-the-art performance.

\section{Methodology}

\begin{figure*}[t] 
  \centering
  \includegraphics[width=0.8\textwidth]{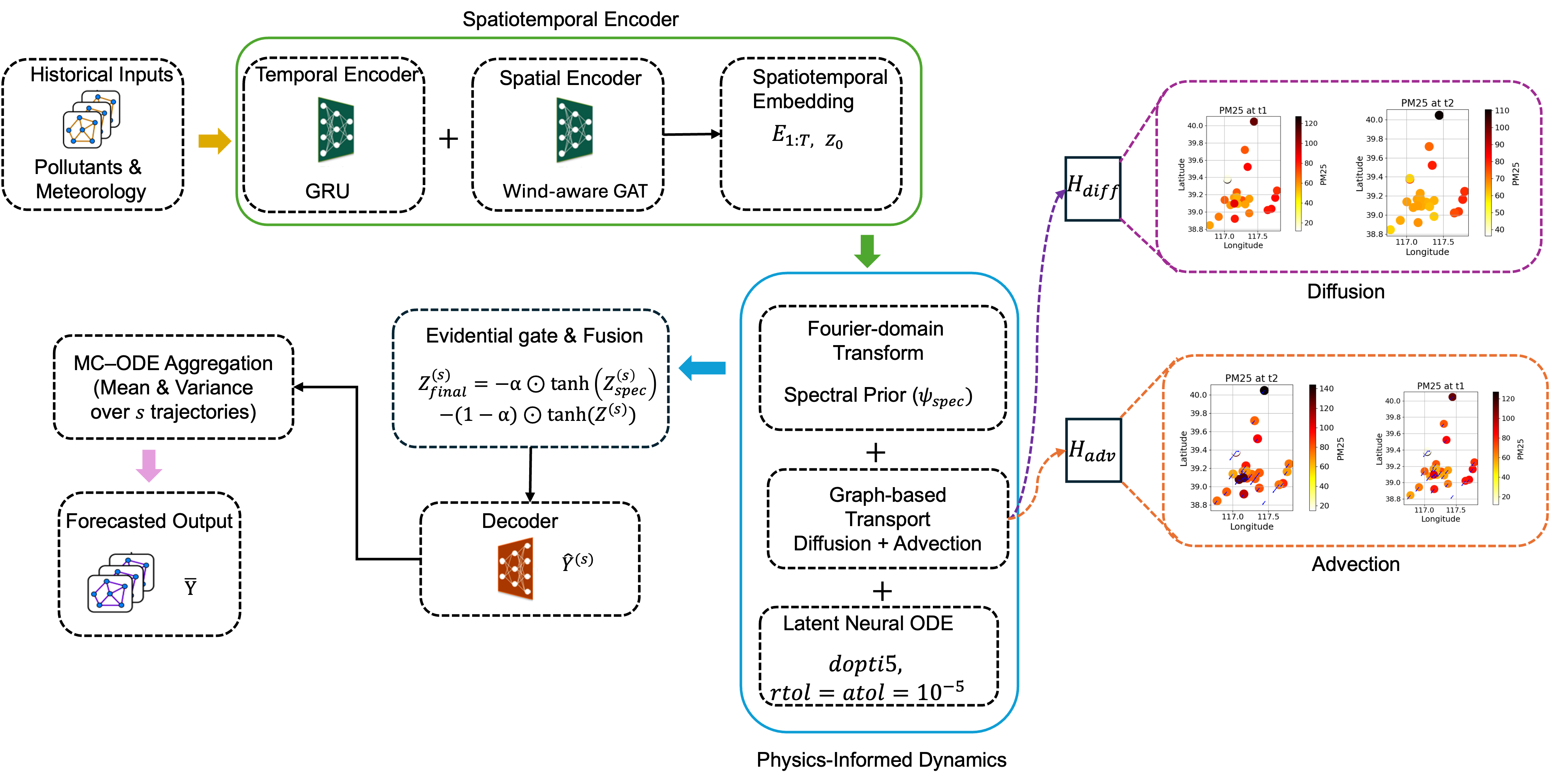}
  \caption{NeuroDDAF framework. Historical pollutant and meteorological sequences are encoded by a GRU and wind-aware GAT, followed by a Fourier-based transport module and latent neural ODE. Physics-guided and residual predictions are fused by an evidential gate with calibrated uncertainty.}
  \label{fig:framework}
\end{figure*}

We propose NeuroDDAF, a physics-informed deep spatio-temporal forecasting framework that integrates neural representation learning with dynamic transport modeling in an open-system setting. NeuroDDAF is designed to address the limitations of purely physics-based and purely data-driven approaches by tightly coupling data-adaptive learning with physically grounded diffusion-advection dynamics. As illustrated in Figure~\ref{fig:framework}, NeuroDDAF consists of four key components:  
(i) a GRU-GAT encoder for spatiotemporal feature extraction,  
(ii) a Fourier-domain diffusion-advection solver with learnable residuals,  
(iii) a wind-aware latent Neural ODE driven by time varying graph operators, and  
(iv) an evidential fusion module that blends physics-guided and data-driven predictions while quantifying uncertainty.

\subsection{Spatiotemporal Encoder: GRU-GAT}
\label{sec:encoder}

Air pollutant dynamics exhibit strong temporal dependencies (e.g., seasonal cycles) as well as spatial coupling driven by atmospheric transport and wind forcing. To capture both effects, NeuroDDAF employs a dual-path spatiotemporal encoder that combines a GRU for temporal modeling with a wind-aware GAT for spatial interaction learning. Given historical observations $\{\mathbf{x}_{t-k}, \dots, \mathbf{x}_t\}$ at each monitoring station, the GRU updates its hidden state as
\begin{equation}
\mathbf{h}_t = \mathrm{GRU}(\mathbf{x}_t, \mathbf{h}_{t-1})
\end{equation}
where the gating mechanism adaptively represents short and long-term temporal evolution. 

In parallel, the monitoring network is represented as a time-varying graph $\mathcal{G}_t = (\mathcal{V}, \mathcal{E}_t)$, where nodes correspond to stations and edges encode meteorological interactions modulated by wind. For node $i$, the GAT computes a spatial embedding
\begin{equation}
\mathbf{z}_i = \sigma\!\left(\sum_{j \in \mathcal{N}(i)} \alpha_{ij}\, \mathbf{W}\mathbf{h}_j \right)
\end{equation}
where $\alpha_{ij}$ are attention coefficients conditioned on connectivity and wind features, $\mathbf{W}$ is a learnable projection matrix, and $\sigma(\cdot)$ denotes a nonlinear activation.

The final spatiotemporal representation is obtained by concatenation,
\begin{equation}
\mathbf{E}_t = \mathrm{Concat}(\mathbf{h}_t, \mathbf{z}_t)
\end{equation}
and serves as conditioning input for the downstream physics-informed modules and the latent Neural ODE.

\subsection{Fourier-based Neural PDE Solver}
\label{sec:pde}
To incorporate physical priors while retaining learning flexibility, we employ a Fourier-domain solver that approximates the diffusion-advection partial differential equation (PDE). The classical transport equation can be expressed as:
\begin{equation}
\frac{\partial c}{\partial t} = D \nabla^{2}c - \mathbf{v} \cdot \nabla c + \mathcal{S}(t)
\end{equation}
where $c$ denotes pollutant concentration, $D$ is the diffusion coefficient, $\mathbf{v}$ represents wind-driven advection, and $\mathcal{S}(t)$ denotes source terms.

We map the concentration field into the spectral domain using the Fourier transform:
\begin{equation}
\hat{c}(\mathbf{k}, t) = \mathcal{F}[c(\mathbf{x}, t)]
\end{equation}
where $\mathbf{k}$ is the frequency vector. In this domain, spatial derivatives reduce to element-wise multiplications, allowing efficient computation of diffusion and advection effects. For example:
\begin{equation}
\mathcal{F}[\nabla^{2}c] = -||\mathbf{k}||^{2}\hat{c}(\mathbf{k}, t)
\end{equation}

While the Fourier operator captures the dominant smooth diffusion-advection behavior, it cannot account for irregular or nonlinear processes (e.g., chemical reactions, sudden emission spikes). To address this, we introduce a learnable residual operator $\mathcal{R}_{\theta}$ in the spectral domain:
\begin{equation}
\frac{\partial \hat{c}}{\partial t} = -D ||\mathbf{k}||^{2}\hat{c} - i(\mathbf{v}\cdot\mathbf{k})\hat{c} + \mathcal{R}_{\theta}(\hat{c}, \mathbf{k}, t)
\end{equation}
where $\mathcal{R}_{\theta}$ is parameterized by a neural network and trained end-to-end. The corrected spectral representation is projected back into the spatial domain via the inverse Fourier transform:
\begin{equation}
c(\mathbf{x}, t+\Delta t) = \mathcal{F}^{-1}\left[\hat{c}(\mathbf{k}, t) + \Delta t \cdot \frac{\partial \hat{c}}{\partial t}\right]
\end{equation}

This Fourier-based solver provides two key advantages:  
(i) it enforces physics-inspired inductive bias, ensuring that learned dynamics remain interpretable; and  
(ii) it enables efficient global updates in $\mathcal{O}(N \log N)$ complexity, making it suitable for real-time forecasting across large spatial grids.

\subsection{Latent Neural ODE with Wind-Aware Dynamics}
\label{sec:ode}

Given the spatiotemporal encoder output $\mathbf{E}_t$ (Sec.~\ref{sec:encoder}), NeuroDDAF evolves latent pollutant representations continuously over time using a Neural Ordinary Differential Equation (ODE). Let $\mathbf{z}(t)\in\mathbb{R}^{N\times d_\ell}$ denote the latent state at time $t$ for $N$ monitoring stations and latent dimension $d_\ell$. The latent temporal dynamics are defined as
\begin{equation}
\frac{\mathrm{d}\mathbf{z}(t)}{\mathrm{d}t}
\;=\;
f_{\theta}\big(\mathbf{z}(t),\ \mathbf{E}_t,\ \mathbf{v}(t),\ \mathcal{G}(t)\big)
\label{eq:neuroddaf-ode}
\end{equation}
where $f_{\theta}$ denotes a learnable vector field modulated by (i) encoder features $\mathbf{E}_t$, (ii) wind fields $\mathbf{v}(t)$, and (iii) a time-varying graph $\mathcal{G}(t)$ encoding diffusion-advection transport processes.

\subsubsection{Graph-based diffusion and advection}
To make the transport dynamics explicit, we represent the monitoring network as a time-varying graph. Each node $v_i$ corresponds to a monitoring site, while dynamic edges connect stations with weights that evolve according to meteorological interactions such as wind speed, direction, and atmospheric mixing. This formulation enables pollutant transport to be expressed through graph operators.

\emph{Diffusion.} Spatial dispersion of pollutants is modeled using the graph Laplacian operator $D\,\mathbf{L}(t)\,\mathbf{c}_t$, where $D$ is the diffusion coefficient, $\mathbf{L}(t)$ denotes the time-varying Laplacian computed from edge weights $\{w_{ij,t}\}$, and $\mathbf{c}_t$ is the pollutant concentration vector. This operator enforces local averaging across connected nodes, capturing smooth spreading behavior over space.

\emph{Advection.} Directional transport, corresponding to wind-driven plume movement, is introduced through a wind-aware operator $\mathbf{M}_A(\mathbf{v}_t)\,\mathbf{c}_t$, where $\mathbf{v}_t$ represents the wind field at time $t$. Attention coefficients $\alpha_{ij,t}$ modulate pollutant transfer along dominant wind directions, enabling anisotropic flow modeling consistent with observed meteorological patterns.

Operationally, the diffusion and advection mechanisms are instantiated using scaled powers of the Laplacian and wind matrices to achieve $K$-hop spatial mixing:
\begin{align}
\mathrm{Diffusion:}\quad
\mathbf{H}_{\mathrm{diff}}
&=\sigma\!\left(\sum_{k=0}^{K-1}\Theta^{(k)}_{\mathrm{d}}\ \mathbf{L}^{\,k}\ \mathbf{z}(t)\right) \\[4pt]
\mathrm{Advection:}\quad
\mathbf{H}_{\mathrm{adv}}
&=\sigma\!\left(\sum_{k=0}^{K-1}\Theta^{(k)}_{\mathrm{a}}\ \mathbf{M}^{\,k}\ \mathbf{z}(t)\right)
\end{align}
and fused through a learnable gate $\boldsymbol{\alpha}\in[0,1]^{N\times d_\ell}$:
\begin{align}
\mathbf{H}_{\mathrm{phys}}
&= \boldsymbol{\alpha}\odot \mathbf{H}_{\mathrm{diff}}
   + (1-\boldsymbol{\alpha})\odot \mathbf{H}_{\mathrm{adv}} \label{eq:gate1} \\[4pt]
\boldsymbol{\alpha}
&= \sigma\!\Big(
   [\,\mathbf{H}_{\mathrm{diff}} \;\Vert\; 
     \mathbf{H}_{\mathrm{adv}} \;\Vert\; 
     \phi(\mathbf{v}(t))\,]\mathbf{W}_{\alpha} 
   + \mathbf{b}_{\alpha}
   \Big)
\label{eq:gate}
\end{align}
where $\sigma(\cdot)$ denotes an elementwise nonlinearity and $\phi(\cdot)$ projects wind features into a latent space. The hyperparameter \texttt{filter\_type} (\texttt{diff}, \texttt{adv}, or \texttt{diff\_adv}) specifies the active transport components, allowing targeted ablation and sensitivity analysis. 

To complement the graph-based formulation, NeuroDDAF incorporates a spectral perspective to capture long-range and smooth pollutant dynamics. A learnable real-valued Fast Fourier Transform (FFT) is applied along the temporal axis of the latent sequence, and the lowest $m{=}8$ frequency modes are modulated with trainable complex-valued weights. This operation compactly models smooth diffusion-advection behavior in the frequency domain while preserving temporal coherence via the inverse FFT. The detailed spectral transport formulation and its training implementation are described in Sec.~\ref{sec:pde}.

The final ODE vector field augments the physics-informed transport operators with encoder-conditioned residual dynamics, enabling NeuroDDAF to model both physically grounded and data-adaptive evolution of latent pollutant states. Formally, the vector field is defined as:
\begin{equation}
\begin{split}
f_{\theta}(\mathbf{z},\mathbf{E}_t,\mathbf{v}(t),\mathcal{G}(t))
&= \underbrace{\mathbf{H}_{\mathrm{phys}}}_{\text{graph-based transport}}
 + \underbrace{\Psi_{\mathrm{spec}}}_{\text{spectral transport}} \\
&\quad + \underbrace{\Phi_{\theta}(\mathbf{z},\mathbf{E}_t)}_{\text{data-driven residual}}
\end{split}
\label{eq:ftheta}
\end{equation}
where $\mathbf{H}_{\mathrm{phys}}$ encodes the graph-based diffusion-advection operators, $\Psi_{\mathrm{spec}}$ represents the Fourier-domain spectral transport, and $\Phi_{\theta}$ captures residual corrections guided by the encoder features $\mathbf{E}_t$.

The latent dynamics in Eq.~\eqref{eq:neuroddaf-ode} are integrated using an adaptive-step Dormand-Prince solver (\texttt{dopri5}) with relative and absolute tolerances set to $\texttt{rtol}=10^{-5}$ and $\texttt{atol}=10^{-5}$, respectively. To enhance robustness and quantify epistemic variability, we draw $S$ trajectories with distinct stochastic realizations (e.g., dropout or perturbed initial conditions):
\begin{equation}
\begin{split}
\{\mathbf{z}^{(s)}(t_{1{:}\tau})\}_{s=1}^{S}
&= \mathrm{ODEINT}\big(\mathbf{z}(t_0), f_{\theta}, \{t_1,\ldots,t_{\tau}\},\\
&\quad \texttt{dopri5},\ \texttt{rtol},\ \texttt{atol}\big),\\
&\quad S = \texttt{n\_traj\_samples}
\end{split}
\end{equation}

We set the sequence length to $T{=}24$, forecast horizon $\tau{=}24$, latent dimension $d_\ell{=}4$, GRU hidden size $64$, $K{=}2$ propagation steps, and $S{=}3$ stochastic trajectories by default.

\subsection{Evidential Fusion and Uncertainty Quantification}
\label{sec:evidential}

NeuroDDAF integrates physics-consistent transport predictions with data-driven residual corrections and explicitly quantifies uncertainty arising from both model variability and data noise. This fusion mechanism ensures reliable forecasting under dynamically varying meteorological conditions. It fuses the latent trajectory obtained from the Neural ODE with the Fourier-filtered latent representation to jointly capture physics-informed and data-driven dynamics. Let $\mathbf{Z}$ denote the latent trajectory produced by the ODE after Monte Carlo (MC) dropout, and let $\mathbf{Z}_{\text{spec}}$ represent the Fourier-filtered output described in Sec.~\ref{sec:pde}. A lightweight gating network processes the concatenated representation $[\mathbf{Z}_{\text{spec}}\Vert \mathbf{Z}\Vert \boldsymbol{\gamma}]$, where $\boldsymbol{\gamma}$ is a random auxiliary tensor that introduces stochastic variation, to generate a gating coefficient $\boldsymbol{\alpha}\in[0,1]$ for adaptive fusion.

\subsubsection{Evidential gate (physics-data fusion)}
For each of the $S$ sampled latent trajectories from the Neural ODE, we fuse the physics-based (spectral) and data-driven (neural) predictions using an evidential gating network:
\begin{align}
\boldsymbol{\alpha}
&=\sigma\!\big(\mathrm{MLP}([\mathbf{Z}_{\text{spec}}\Vert \mathbf{Z}\Vert \boldsymbol{\gamma}])\big),
\label{eq:fusion_alpha} \\[4pt]
\mathbf{Z}_{\text{final}}^{(s)}
&= -\,\boldsymbol{\alpha}\odot \tanh(\mathbf{Z}_{\text{spec}}^{(s)})
  - (1-\boldsymbol{\alpha})\odot \tanh(\mathbf{Z}^{(s)}),
\label{eq:fusion_final}
\end{align}
where $\mathbf{Z}_{\text{spec}}^{(s)}$ and $\mathbf{Z}^{(s)}$ denote the spectral and ODE outputs for trajectory $s$. The fused latent fields $\{\mathbf{Z}_{\text{final}}^{(s)}\}_{s=1}^{S}$ are subsequently decoded into predictions and used to compute ensemble-based uncertainty estimates.
The evidential gate adaptively balances physics-informed and neural residual predictions at each spatio-temporal step. When the Fourier-based physics prior captures dominant dynamics (e.g., wind-driven advection or diffusion), the gate increases reliance on the physics branch via higher $\boldsymbol{\alpha}$. Conversely, under irregular, nonlinear, or data-dominated regimes, the neural branch contributes more strongly. This mechanism yields robust, uncertainty-aware fusion that preserves physical interpretability while maintaining flexibility for complex, data-driven corrections.

\subsubsection{Ensemble mean and variance (MC-ODE)}
We draw $S$ trajectories with different stochastic realizations (e.g., dropout or initial noise), where $S$ is controlled by \texttt{n\_traj\_samples} (default $S{=}1$; $S{=}3$ in experiments). From these $S$ fused trajectories, we compute the predictive mean and total variance using the law of total variance:
\begin{align}
\bar{\mathbf{Y}}
&=\frac{1}{S}\sum_{s=1}^{S}\hat{\mathbf{Y}}^{(s)} \\[4pt]
\mathbf{V}_{\text{tot}}
&=\underbrace{\frac{1}{S}\sum_{s=1}^{S}\boldsymbol{\Sigma}^{(s)}_{\text{ale}}}_{\text{aleatoric (data) uncertainty}}
+ \underbrace{\frac{1}{S}\sum_{s=1}^{S}\big(\hat{\mathbf{Y}}^{(s)}-\bar{\mathbf{Y}}\big)^{\!2}}_{\text{epistemic (model) uncertainty}}
\end{align}
where $\boldsymbol{\Sigma}^{(s)}_{\text{ale}}$ denotes the aleatoric variance produced by a heteroscedastic head when enabled; otherwise, the first term is omitted and $\mathbf{V}_{\text{tot}}$ reduces to the empirical trajectory variance. Prediction intervals are reported as $\bar{\mathbf{Y}} \ \pm\ z_{1-\alpha/2}\sqrt{\mathbf{V}_{\text{tot}}}$.

\subsubsection{Evidential regression}
When the evidential head is activated, each output dimension parameterizes a conjugate uncertainty model (e.g., Normal-Inverse-Gamma for scalar regression), enabling closed-form posterior updates and interpretable uncertainty estimation. The network predicts evidence parameters $\{\mu,\lambda,\alpha,\beta\}$ constrained by $\alpha{>}1$ and $\beta{>}0$. The predictive likelihood follows a Student-$t$ distribution, and the evidential loss is defined as
\begin{equation}
\mathcal{L}_{\text{evidential}}
= \underbrace{-\log p\big(\mathbf{Y}\;\big|\;\mu,\lambda,\alpha,\beta\big)}_{\text{negative log-evidence}}
+ \lambda_{\mathrm{reg}}\cdot \underbrace{\mathcal{R}\big(\mu,\lambda,\alpha,\beta\ ;\ \mathbf{Y}\big)}_{\text{evidence regularizer}}
\end{equation}
which discourages unwarranted certainty by penalizing excessive evidence when residual errors are large.

\begin{algorithm}[t]
\footnotesize
\caption{\textbf{NeuroDDAF: End-to-End Training and Inference Pipeline}}
\label{alg:neuroddaf}
\begin{algorithmic}[1]
  \Require Historical sequences $\{\mathbf{X}_{t-T+1},\dots,\mathbf{X}_{t}\}$; targets $\mathbf{Y}_{t+1:t+\tau}$; dynamic graph $\mathcal{G}=(\mathcal{V},\mathcal{E},\mathbf{A},\texttt{edge\_index},\texttt{edge\_attr})$; horizon $\tau$; samples $S$
  \Ensure Forecasts $\hat{\mathbf{Y}}_{t+1:t+\tau}$ and prediction intervals
  \State \textbf{Init:} $\Theta=\{\Theta_{\text{enc}},\Theta_{\text{ode}},\Theta_{\text{fourier}},\Theta_{\text{gate}},\Theta_{\text{dec}}\}$; Adam; \texttt{dopri5} with $\texttt{rtol}=\texttt{atol}=10^{-5}$

  \For{$e \gets 1$ to $E$}
    \ForAll{batches $(\mathbf{X}_b,\mathbf{Y}_b)$}
      \State \textbf{Step 1: Encoder \& $z_0$ sampling}
      \State Split target channel (e.g., PM$_{2.5}$) and wind variables from $\mathbf{X}_b$
      \State $\mathbf{H}_{1:T} \gets \mathrm{GRU}(\text{target channel})$
      \State $\mathbf{E}_T \gets \mathrm{GAT}(\mathbf{H}_T,\texttt{edge\_index})$
      \State $(\boldsymbol{\mu}_{z_0},\boldsymbol{\sigma}_{z_0}) \gets \mathrm{MLP}(\mathbf{E}_T)$
      \For{$s \gets 1$ to $S$}
        \State Sample $\mathbf{z}^{(s)}(t_0) \gets \boldsymbol{\mu}_{z_0} + \boldsymbol{\sigma}_{z_0}\odot \varepsilon^{(s)}$, \ $\varepsilon^{(s)}\sim\mathcal{N}(0,\mathbf{I})$
      \EndFor

      \State \textbf{Step 2: Latent ODE evolution (physics-aware)}
      \State Build $f_{\theta}$ using $(\mathbf{A},\texttt{edge\_index},\texttt{edge\_attr},\text{wind})$
        \For{$s \gets 1$ to $S$}
          \State $\mathbf{Z}^{(s)} \gets \mathrm{ODESolve}\big(f_{\theta},\mathbf{z}^{(s)}(t_0),\{t_1,\ldots,t_{\tau}\};$
          \State \hspace{1.5em}$\texttt{dopri5},\texttt{rtol},\texttt{atol}\big)$
        \EndFor

      \State \textbf{Step 3: Spectral transport \& evidential fusion}
      \For{$s \gets 1$ to $S$}
        \State $\tilde{\mathbf{Z}}^{(s)} \gets \mathrm{Dropout}(\mathbf{Z}^{(s)})$
        \State $\mathbf{Z}^{(s)}_{\text{spec}} \gets \mathrm{FourierPDE}\!\left(\tilde{\mathbf{Z}}^{(s)}\right)$ \Comment{lowest temporal modes}
        \State Sample $\boldsymbol{\gamma}^{(s)} \sim \mathcal{U}(0,1)$ with shape of $\mathbf{Z}^{(s)}_{\text{spec}}$
        \State $\boldsymbol{\alpha}^{(s)} \gets \sigma\!\Big(\mathrm{MLP}\big([\mathbf{Z}^{(s)}_{\text{spec}}\Vert \tilde{\mathbf{Z}}^{(s)}\Vert \boldsymbol{\gamma}^{(s)}]\big)\Big)$
        \State $\mathbf{Z}^{(s)}_{\text{final}} \gets -\boldsymbol{\alpha}^{(s)}\odot \tanh(\mathbf{Z}^{(s)}_{\text{spec}}) - (1-\boldsymbol{\alpha}^{(s)})\odot \tanh(\tilde{\mathbf{Z}}^{(s)})$
      \EndFor

      \State \textbf{Step 4: Decoding}
      \For{$s \gets 1$ to $S$}
        \State $\hat{\mathbf{Y}}^{(s)} \gets \mathrm{Decoder}\!\left(\mathbf{Z}^{(s)}_{\text{final}}\right)$
      \EndFor

      \State \textbf{Step 5: Aggregation \& uncertainty}
      \State $\bar{\mathbf{Y}} \gets \frac{1}{S}\sum_{s=1}^{S}\hat{\mathbf{Y}}^{(s)}$
      \State $\mathbf{V}_{\text{tot}} \gets \frac{1}{S}\sum_{s=1}^{S}\boldsymbol{\Sigma}^{(s)}_{\text{ale}} \;+\; \frac{1}{S}\sum_{s=1}^{S}\big(\hat{\mathbf{Y}}^{(s)}-\bar{\mathbf{Y}}\big)^2$

      \State \textbf{Step 6: Loss \& update}
      \State $\mathcal{L}_{\text{forecast}} \gets \mathrm{MAE/RMSE}(\bar{\mathbf{Y}},\mathbf{Y}_b)$
      \State $\mathcal{L}_{\text{PDE}} \gets \|P_h[\bar{\mathbf{Y}}]\|_2^2$
      \State $\mathcal{L}_{\text{evidential}} \gets \mathrm{EvidentialNLL}\!\left(\{\hat{\mathbf{Y}}^{(s)}\}_{s=1}^{S}\right)$
      \State $\mathcal{L} \gets \mathcal{L}_{\text{forecast}} + \lambda_{\text{phys}}\mathcal{L}_{\text{PDE}} + \lambda_{\text{unc}}\mathcal{L}_{\text{evidential}}$
      \State $\Theta \gets \mathrm{AdamUpdate}(\Theta,\nabla_{\Theta}\mathcal{L})$
    \EndFor
  \EndFor

  \State \textbf{Inference:} For unseen $\mathbf{X}_{t-T+1:t}$, repeat Steps 1--5 (no gradients) to obtain $\bar{\mathbf{Y}}_{t+1:t+\tau}$ and prediction intervals from $\mathbf{V}_{\text{tot}}$.
\end{algorithmic}
\end{algorithm}

\subsection{Training Objective}
\label{sec:training}

The overall optimization objective of NeuroDDAF jointly enforces accurate forecasting, physical consistency, and calibrated uncertainty estimation. The total loss is defined as:
\begin{equation}
\mathcal{L}
= \mathcal{L}_{\text{forecast}}(\bar{\mathbf{Y}}, \mathbf{Y})
+ \lambda_{\text{phys}}\, \mathcal{L}_{\text{PDE}}
+ \lambda_{\text{unc}}\, \mathcal{L}_{\text{evidential}}
\end{equation}
where $\mathcal{L}_{\text{forecast}}$ corresponds to the forecasting error. The physics-consistency term $\mathcal{L}_{\text{PDE}}$ penalizes residuals of the discretized transport operator derived from the Fourier-domain prior (Sec.~\ref{sec:pde}), ensuring that predictions remain consistent with diffusion-advection dynamics. The evidential term $\mathcal{L}_{\text{evidential}}$ is applied only when the uncertainty head is enabled, promoting calibrated and interpretable uncertainty estimates. The end-to-end procedure is summarized in Algorithm~\ref{alg:neuroddaf}.

\section{Experiment}

\subsection{Experiment Setting}

\textbf{Datasets.}  
We evaluate NeuroDDAF and state-of-the-art (SOTA) baselines on four real-world air quality datasets: Beijing\footnote{\url{https://dataverse.harvard.edu/dataverse/whw195009}} (35 stations, 2017-2018), Shenzhen\footnote{\url{https://www.microsoft.com/en-us/research/project/urban-air/}\label{fn:urban-air}} (11 stations, 2014-2015), Tianjin\footref{fn:urban-air} (27 stations, 2014-2015), and Ancona\footnote{\url{https://zenodo.org/records/11220965}} (21 stations, 2021-2023). Following prior work~\cite{hettige2024airphynet}, we forecast PM$_{2.5}$ concentrations using meteorological factors (wind speed and wind direction) as covariates. Missing values are imputed using a 24-hour moving mean. Data are split into 70\%/10\%/20\% for training/validation/testing, with a 3-hour resolution. The preceding 72 hours (24 steps) are used to predict 1-, 2-, and 3-day horizons. All models are trained on NVIDIA A100 GPUs using the Adam optimizer (batch size $32$, initial learning rate $\eta_0=5\times10^{-4}$ with step decay). Training and testing strictly follow the AirPhyNet setup~\cite{hettige2024airphynet}. 

\textbf{Baselines.}  
We compare against eleven SOTA methods: (1) statistical HA, VAR; (2) neural ODE-based Latent-ODE~\cite{chen2018neural}, ODE-RNN~\cite{rubanova2019latent}, ODE-LSTM~\cite{lechner2020learning}; (3) spatio-temporal deep learning DCRNN~\cite{li2017diffusion}, STGCN~\cite{yu2017spatio}, ASTGCN~\cite{guo2019attention}, GTS~\cite{shang2021discrete}, MTSF-DG~\cite{zhao2023multiple}, PM2.5-GNN~\cite{wang2020pm2}, AirFormer~\cite{liang2023airformer}, and (4) physics-informed AirPhyNet~\cite{hettige2024airphynet}. Further experimental configurations, hyperparameter settings, and theoretical assumptions (well-posedness, stability, uncertainty guarantees) are provided in Appendices~A–C.

\subsection{Performance Comparison}

\begin{table*}[t]
\caption{Performance comparison on the Beijing dataset in terms of MAE and RMSE for 1-day, 2-day, and 3-day forecasts. The best results are highlighted in \textbf{bold}, while the second-best results are \underline{underlined}.}
\vskip 0.15in
\begin{center}
\footnotesize
\begin{sc}
\begin{tabular}{l|cc|cc|cc}
\toprule
\multirow{2}{*}{Model} & \multicolumn{6}{c}{Beijing data} \\ \cline{2-7}
 & \multicolumn{2}{c|}{1-day} & \multicolumn{2}{c|}{2-day} & \multicolumn{2}{c}{3-day} \\ \cline{2-7}
 & MAE & RMSE & MAE & RMSE & MAE & RMSE \\ \midrule
HA & 38.37 & 45.8 & 83.91 & 95.56 & 50.58 & 101.51 \\
VAR & 60.10 & 102.92 & 60.44 & 103.02 & 60.64 & 103.07 \\
\midrule
DCRNN & 35.99 & 52.55 & 49.66 & 67.50 & 57.01 & 74.67 \\
STGCN & 33.70 & 49.16 & 38.93 & 54.98 & 43.93 & 56.57 \\
GMAN & 50.62 & 66.05 & 50.73 & 66.07 & 50.69 & 65.87 \\
GTS & 34.99 & 51.45 & 54.18 & 71.87 & 73.50 & 89.59 \\
PM25GNN & 50.94 & 65.87 & 48.81 & 65.64 & 51.51 & 66.55\\
AirFormer & \underline{29.62} & 46.49 & 38.43 & 56.52 & 43.39 & 58.68\\
\midrule
LatentODE & 44.83 & 53.96 & 45.95 & 55.44 & 47.14 & 57.39\\
ODE-LSTM & 46.19 & 57.56 & 49.18 & 62.39 & 51.45 & 63.66\\
\midrule
AirPhyNet & \textbf{29.47} & \underline{42.54} & 37.26 & 49.25 & 43.21 & 54.11\\
NeuroDDAF (GRU) & 31.53 & 43.44 & \underline{36.72} & \underline{48.06} & \underline{39.98} & \underline{50.44}\\
NeuroDDAF (LSTM) & 31.90 & 44.09 & 37.24 & 48.80 & 40.43 & 51.06\\
NeuroDDAF (TCN) & 41.30 & 53.05 & 43.88 & 56.53 & 45.05 & 55.78 \\
NeuroDDAF (GPT2) & 48.54 & 59.94 & 49.65 & 62.37 & 50.05 & 61.16\\
NeuroDDAF (Transformer) & 42.15 & 53.90 & 43.59 & 56.25 & 44.83 & 55.70 \\
\textbf{NeuroDDAF (GRU-GAT)} & 29.97 & \textbf{41.63} & \textbf{35.42} & \textbf{46.35} & \textbf{39.16} & \textbf{48.88}\\
\midrule
Improvement vs AirPhyNet & – & \textbf{2.1\%} & \textbf{4.9\%} & \textbf{5.9\%} & \textbf{9.4\%} & \textbf{9.7\%} \\
\bottomrule
\end{tabular}
\end{sc}
\end{center}
\label{tab:beijing_data}
\vskip -0.1in
\end{table*}

Table~\ref{tab:beijing_data} presents the MAE and RMSE of all baseline and proposed models on the Beijing dataset for 1-day, 2-day, and 3-day forecasts. The traditional baselines, HA and VAR, show relatively high errors, particularly for multi-day horizons, highlighting their inability to capture the complex spatiotemporal dependencies in air quality forecasting. In contrast, classical deep learning models such as DCRNN, STGCN, GMAN, GTS, and PM25GNN achieve substantial improvements over HA and VAR. Among them, STGCN demonstrates consistently strong performance in the 1-day and 2-day settings, while GMAN and PM25GNN perform less competitively on this dataset. AirFormer also achieves strong performance, with its 1-day MAE (29.62~$\mu$g/m$^3$) already being competitive with more specialized architectures. Meanwhile, continuous-time models such as LatentODE and ODE-LSTM provide only limited gains and generally lag behind discrete-time methods, suggesting that they struggle to capture fine-grained dynamics in this context. AirPhyNet emerges as a competitive baseline, offering lower RMSE values compared to earlier deep learning approaches, and serves as a strong point of comparison for evaluating newer architectures.

Within the NeuroDDAF family, the GRU- and LSTM-based variants stand out as consistently strong performers, surpassing most competing baselines. In particular, the GRU variant achieves second-best results across forecast horizons, while Transformer- and GPT2-based variants underperform, indicating that temporal alignment and inductive biases play a crucial role in performance. Most notably, NeuroDDAF (GRU-GAT) delivers the best results overall, achieving the lowest RMSE for 1-day forecasts (41.63~$\mu$g/m$^3$) and the lowest MAE and RMSE across all horizons. Compared to AirPhyNet, it reduces RMSE by 2.1\%, 5.9\%, and 9.7\% for 1-day, 2-day, and 3-day horizons, respectively.

Although these improvements may appear numerically modest, they can translate into meaningful real-world impacts. For example, reducing average PM$_{2.5}$ forecast error by even 1~$\mu$g/m$^3$ can shift predicted concentrations across regulatory thresholds (e.g., WHO 24-hour exposure limit), directly influencing public health advisories and mitigation policies. Thus, even a 1-2\% reduction in MAE or RMSE is practically valuable, underscoring the importance of consistent performance gains delivered by NeuroDDAF across horizons.

\subsection{Ablation Study}

\begin{table*}[t]
\caption{Performance (MAE and RMSE) for 1-/2-/3-day forecasts across four datasets. AirPhyNet vs. NeuroDDAF encoder variants. The best results are highlighted in \textbf{bold}, while the second-best results are \underline{underlined}.}
\label{tab:models_as_columns_grouped_encoders}
\vskip 0.10in
\begin{center}
\footnotesize
\begin{sc}
\resizebox{\textwidth}{!}{%
\begin{tabular}{c|c*{7}{|cc}}
\toprule
\multirow{3}{*}{\textbf{Dataset}} & \multirow{3}{*}{\textbf{Day}}
& \multicolumn{2}{c|}{\textbf{AirPhyNet}}
& \multicolumn{12}{c}{\textbf{NeuroDDAF (Encoder)}} \\ \cline{5-16}
& &
\multicolumn{2}{c|}{} 
& \multicolumn{2}{c|}{\textbf{GRU-GAT}}
& \multicolumn{2}{c|}{\textbf{GRU}}
& \multicolumn{2}{c|}{\textbf{LSTM}}
& \multicolumn{2}{c|}{\textbf{TCN}}
& \multicolumn{2}{c|}{\textbf{GPT2}}
& \multicolumn{2}{c}{\textbf{Transformer}} \\ \cline{3-16}
& &
\textbf{MAE} & \textbf{RMSE}
& \textbf{MAE} & \textbf{RMSE}
& \textbf{MAE} & \textbf{RMSE}
& \textbf{MAE} & \textbf{RMSE}
& \textbf{MAE} & \textbf{RMSE}
& \textbf{MAE} & \textbf{RMSE}
& \textbf{MAE} & \textbf{RMSE} \\ \midrule
\multirow{3}{*}{Shenzen}
& 1-day & \textbf{10.55} & \textbf{13.99} & 11.22 & 14.85 & 11.34 & 15.00 & 11.30 & 14.96 & 10.99 & 14.59 & \underline{10.98} & \underline{14.58} & 14.26 & 17.44 \\
& 2-day & 12.46 & \textbf{16.04} & 12.73 & 16.51 & 12.86 & 16.72 & 12.74 & 16.56 & \underline{12.30} & 16.08 & \textbf{12.27} & \underline{16.05} & 14.58 & 17.85  \\
& 3-day & 13.12 & 16.50 & 13.03 & 16.55 & 13.12 & 16.62 & 13.03 & 16.60 & \textbf{12.69} & \textbf{16.13} & \underline{12.72} & \underline{16.15} & 15.05 & 18.05  \\ \midrule

\multirow{3}{*}{Tianjin}
& 1-day & 29.13 & 38.34 & \textbf{28.29} & \textbf{36.95} & 29.39 & 37.90 & \underline{28.60} & \underline{37.37} & 28.79 & 37.80 & 29.76 & 38.31 & 29.94 & 38.64  \\
& 2-day & 35.30 & 45.38 & \textbf{34.09} & \textbf{43.48} & 35.26 & 44.44 & \underline{34.46} & \underline{44.04} & 35.14 & 44.93 & 35.47 & 44.77 & 35.44 & 44.92  \\
& 3-day & 39.08 & 49.38 & \textbf{37.27} & \textbf{46.79} & 38.64 & 47.98 & \underline{38.07} & \underline{47.79} & 38.44 & 48.39 & 38.95 & 48.47 & 39.32 & 48.93 \\ \midrule

\multirow{3}{*}{Ancoda}
& 1-day & \underline{2.63} & \underline{3.73} & \textbf{2.61} & \textbf{3.70} & 2.65 & 3.74 & 2.97 & 4.12 & 3.07 & 4.22 & 3.25 & 4.34 & 4.10 & 5.08  \\
& 2-day & \underline{3.03} & \underline{4.13} & \textbf{3.01} & \textbf{4.11} & 3.03 & 4.13 & 3.20 & 4.38 & 3.41 & 4.56 & 3.59 & 4.70 & 4.27 & 5.26  \\
& 3-day & \underline{3.18} & \underline{4.21} & \textbf{3.14} & \textbf{4.18} & 3.19 & 4.23 & 3.24 & 4.34 & 3.62 & 4.75 & 3.71 & 4.77 & 4.37 & 5.37  \\ \midrule
\multicolumn{2}{c|}{Top Count} & \multicolumn{2}{c|}{3} & \multicolumn{2}{c|}{12} & \multicolumn{2}{c|}{0} & \multicolumn{2}{c|}{0} & \multicolumn{2}{c|}{2} & \multicolumn{2}{c|}{1} & \multicolumn{2}{c}{0}\\
\bottomrule
\end{tabular}
} 
\end{sc}

\end{center}
\vskip -0.1in
\end{table*}

\textbf{Encoder Agnostic Performance.} 
Table~\ref{tab:models_as_columns_grouped_encoders} compares AirPhyNet with NeuroDDAF under different encoder backbones across Shenzhen, Tianjin, and Ancona. Overall, NeuroDDAF remains competitive or improves over AirPhyNet across most horizons, demonstrating that the framework is encoder-agnostic. Among NeuroDDAF variants, GRU-GAT achieves the strongest overall results (highest top-count), while GRU/LSTM variants are close behind, indicating that temporal modeling dominates in these datasets and graph-based spatial attention provides incremental but consistent gains. Transformer-based encoders generally underperform, suggesting that additional capacity does not translate into improved forecasting under the current data regime.
\begin{figure*}[t] 
  \centering
  \includegraphics[width=.80\textwidth]{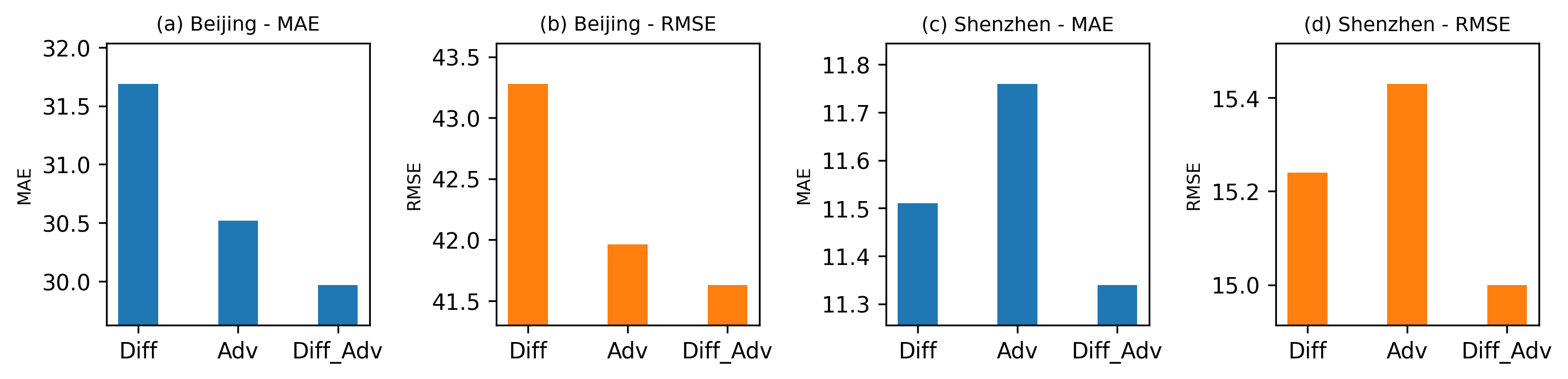}
  \caption{Effect of Physical knowledge in NeuroDDAF, (a) shows MAE in Beijing, (b) shows RMSE in Beijing,
  (c) shows MAE in Shenzhen, and (d) shows RMSE in Shenzhen.}
  \label{fig:gru-gat-mae-rmse}
\end{figure*}

\textbf{Effect of Physical Knowledge.} To further evaluate the impact of incorporating physical knowledge, we compared our proposed model NeuroDAAF with two simplified variants: (a) Diff, which only considers the diffusion process, and (b) Adv, which only considers the advection process. We conducted experiments on both the Beijing and Shenzhen datasets. As shown in Figure~\ref{fig:gru-gat-mae-rmse}, NeuroDAAF consistently achieves lower MAE and RMSE values than either the Diff or Adv variants. In Beijing, incorporating both diffusion and advection significantly reduces error, while in Shenzhen, NeuroDAAF similarly demonstrates superior performance across both metrics. These results clearly highlight the benefit of integrating physical knowledge into the deep learning framework, as the joint modeling of diffusion and advection processes leads to more accurate and robust air quality forecasts.

\subsection{Computation Cost and Runtime Analysis}
\label{sec:runtime}

\begin{figure*}[!ht]
    \centering
    \includegraphics[width=0.75\textwidth]{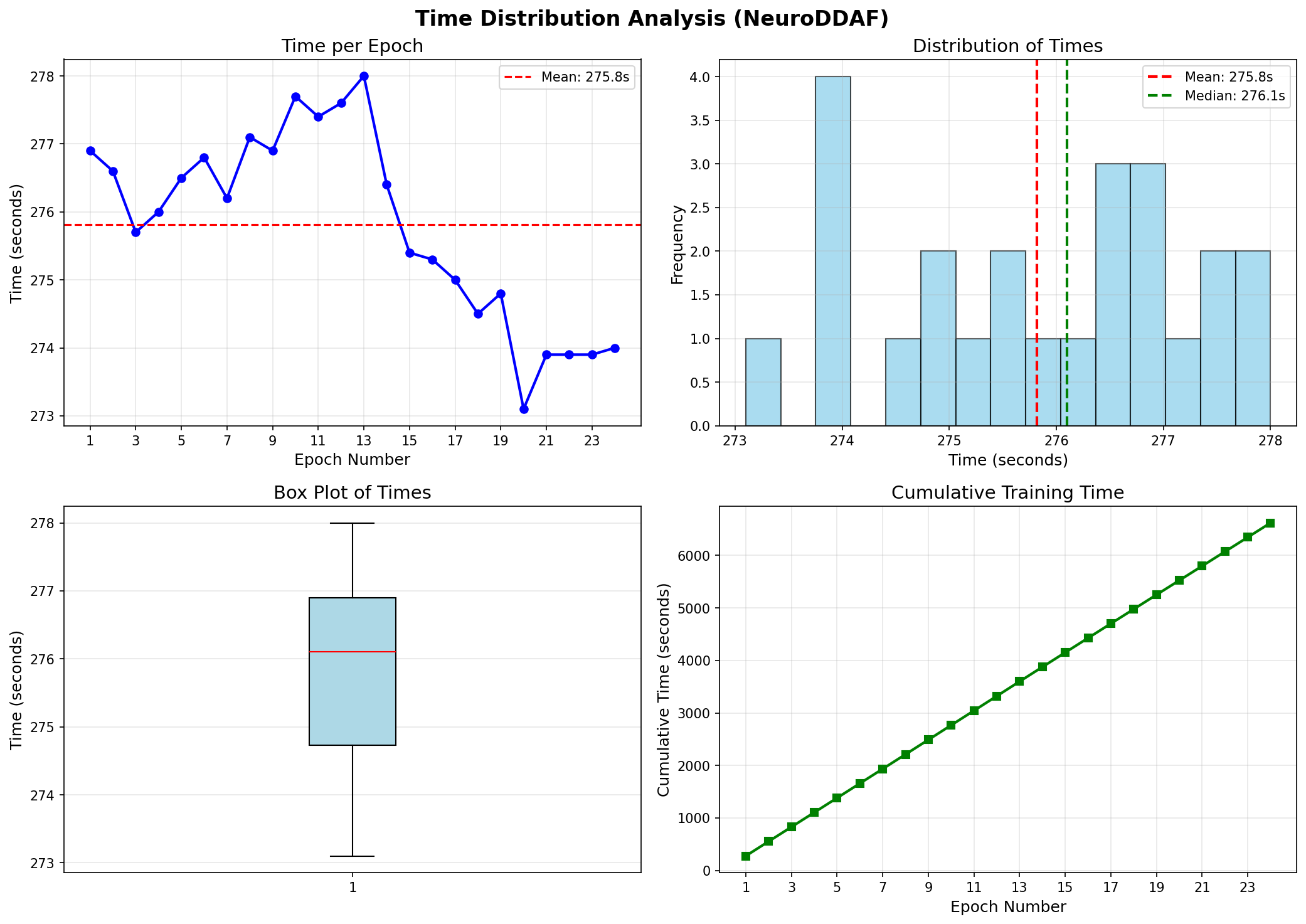}
    \caption{Computation-time analysis for NeuroDDAF on Beijing. Epoch time remains stable across training (top-left) and forms a tight distribution (top-right), with consistent spread in the box plot (bottom-left) and near-linear cumulative training time (bottom-right).}
    \label{fig:time_analysis}
\end{figure*}

In addition to predictive accuracy, we measure empirical runtime and model size using the Beijing split. Table~\ref{tab:runtime} summarizes parameters, batches/epoch, average epoch time, and early-stopping epochs for each encoder variant. A key observation is that the per-epoch time is stable across encoders despite large differences in parameter count. This indicates that the dominant cost comes from ODE integration and transport operators, rather than the encoder backbone. Figure~\ref{fig:time_analysis} further supports this: epoch times cluster tightly around a mean of $\approx$275.8~s with limited spread. Notably, NeuroDDAF (GRU-GAT) is below the overall average in both parameter count ($\sim$75k vs.\ $\sim$646k average) and epoch time ($\sim$276~s vs.\ $\sim$286~s average), while achieving the best forecasting performance. Therefore, differences in total wall-clock training time are driven primarily by early stopping, not by encoder computation.

\begin{table}[t]
\centering
\caption{Empirical runtime and model size on the Beijing dataset (batch size 32, fixed solver tolerances, identical hardware). Avg.\ epoch time in seconds.}
\label{tab:runtime}
\footnotesize
\setlength{\tabcolsep}{3.5pt}
\renewcommand{\arraystretch}{1.05}
\resizebox{\columnwidth}{!}{%
\begin{tabular}{lrrrr}
\toprule
\textbf{Encoder} & \textbf{Params} & \textbf{Batches/epoch} & \textbf{Epoch time (s)} & \textbf{Stop (ep.)} \\
\midrule
GRU-GAT      & 75,594     & 69 & $\sim$276 & 24 \\
GRU          & 71,306     & 69 & $\sim$270 & 25 \\
LSTM         & 75,594     & 69 & $\sim$271 & 23 \\
TCN          & 84,362     & 69 & $\sim$272 & 42 \\
Transformer  & 125,834    & 69 & $\sim$273 & 40 \\
GPT-2        & 3,440,970  & 69 & $\sim$356 & 22 \\
\midrule
Average      & $\sim$645,610 & 69 & $\sim$286 & $\sim$29 \\
\bottomrule
\end{tabular}%
}
\end{table}

\subsection{Case Study: Wind Effects on PM$_{2.5}$ Concentration}

\begin{figure*}[t]
    \centering
    \includegraphics[width=.80\linewidth]{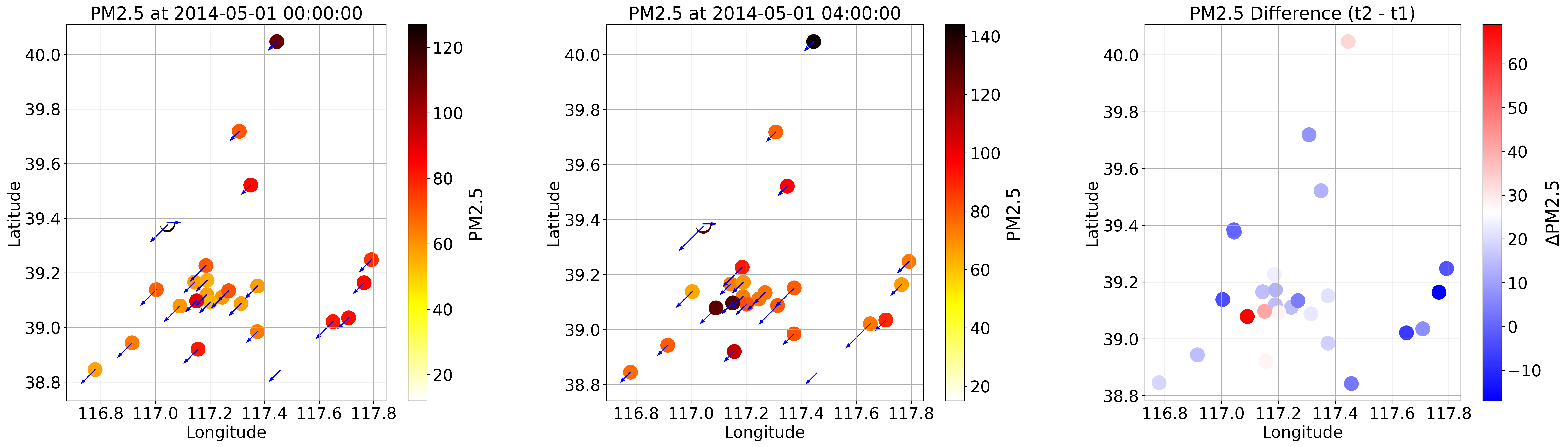}
    \caption{PM$_{2.5}$ spatial distribution in Tianjin on May 1, 2014 at 00:00 (left) and 04:00 (middle) with wind vectors, and their difference (right). Wind-driven transport contributes to localized increases and decreases in concentration.}
    \label{fig:pm25_case}
\end{figure*}

To further illustrate how atmospheric dynamics influence air pollution, we analyze PM$_{2.5}$ variations in Tianjin on May 1, 2014. Figure~\ref{fig:pm25_case} (left and middle) shows the spatial distribution of PM$_{2.5}$ concentrations at 00:00 and 04:00, while the right panel displays the difference between these snapshots. Although absolute levels remained elevated across multiple stations, the difference map highlights localized increases and decreases that align partially with prevailing wind vectors. In general, stations situated downwind of higher-emission regions tended to experience elevated concentrations, while some upwind stations exhibited reductions, consistent with wind-driven pollutant transport. 

At the same time, the pattern is not uniform across the network. A sharp local increase at one station may be linked to proximity to an industrial area rather than solely wind dynamics, and several other stations show decreases even in regions not clearly aligned with the wind field. This reflects the heterogeneous nature of air pollution, where wind plays a dominant but not exclusive role. Overall, this case study illustrates how wind direction and speed contribute to pollutant redistribution, while also acknowledging that local sources and site-specific conditions can complicate the observed response. A geospatial visualization with station coordinates and an OpenStreetMap basemap, along with a regime-based classification of transport dynamics, is provided in Appendix~D 

\section{ Conclusion and Future Work}

In this paper, we introduced NeuroDDAF, a physics-informed deep learning framework that unifies pollutant transport dynamics with neural representation learning in an open-system setting. By combining a GRU-Graph Attention encoder, a Fourier-domain PDE solver with neural residuals, a latent neural ODE with dynamic wind coupling, and an evidential fusion mechanism, NeuroDDAF achieves a principled balance between physical interpretability and predictive accuracy. Extensive experiments on four large-scale urban datasets demonstrate that our framework consistently outperforms state-of-the-art baselines, including AirPhyNet, across multiple lead times. Beyond improved error metrics, NeuroDDAF provides robust uncertainty calibration and captures physically meaningful pollutant transport dynamics, as confirmed through case studies. These results establish NeuroDDAF as a reliable and interpretable approach for next-generation air quality forecasting.

Despite these contributions, several directions remain open for future research. The framework could be extended to incorporate additional physical mechanisms, such as chemical transformations or secondary pollutant formation, to enhance realism. Moreover, integrating real-time satellite, sensor, and emission data streams may enable adaptive forecasting under rapidly changing conditions.




\balance

\bibliographystyle{IEEEtran.bst}
\bibliography{./ref/longforms,./ref/references}

\end{document}


\title{NeuroDDAF: Neural Dynamic Diffusion-Advection Fields with Evidential Fusion for Air Quality Forecasting}

\maketitle

\appendices

\section{Experimental Settings}
\label{sec:experimental-settings}
  
We denote the historical input tensor as $\mathbf{X} \in \mathbb{R}^{T \times N \times F}$, containing $T$ time steps of $F$ features (pollutant concentrations and meteorological variables) across $N$ monitoring stations. The target tensor is $\mathbf{Y} \in \mathbb{R}^{\tau \times N \times d'}$, representing $\tau$-step forecasts of $d'$ target pollutants (e.g., PM$_{2.5}$). Latent states are $\mathbf{z}(t) \in \mathbb{R}^{N \times d_\ell}$, wind features are $\mathbf{v}(t) \in \mathbb{R}^{N \times d_w}$, and graph operators are $\mathbf{L}(t), \mathbf{M}_A(\mathbf{v}_t) \in \mathbb{R}^{N \times N}$.  

Performance is assessed using Mean Absolute Error (MAE) and Root Mean Squared Error (RMSE). For predictions $\hat{y}_{t,n}$ and ground truth $y_{t,n}$ across forecast horizon $t=1,\dots,\tau$ and stations $n=1,\dots,N$, the metrics are defined as:
\begin{align}
\text{MAE} &= \frac{1}{\tau N} \sum_{t=1}^{\tau} \sum_{n=1}^{N} \left| y_{t,n} - \hat{y}_{t,n} \right|, \\
\text{RMSE} &= \sqrt{ \frac{1}{\tau N} \sum_{t=1}^{\tau} \sum_{n=1}^{N} \left( y_{t,n} - \hat{y}_{t,n} \right)^2 }.
\end{align}
Lower values indicate better predictive accuracy. All metrics are reported in $\mu$g/m$^3$ for PM$_{2.5}$ concentrations. The default hyperparameters and notational conventions used throughout the paper are summarized in Table~\ref{tab:configuration}.

\begin{table*}[h!]
\caption{Default hyperparameters and symbols used in experiments and theoretical analysis.}
\vskip 0.15in
\begin{center}
\footnotesize
\begin{sc}
\begin{tabular}{l l}
\toprule
\textbf{Quantity} & \textbf{Symbol / Setting} \\
\midrule
Sequence length & $T=24$ (72 hours at 3-hour resolution) \\
Forecast horizon & $\tau=24$ (3-day forecast)\\
Stations / features & $N$ (dataset-specific), $F$ (pollutants+meteo)\\
Output channels & $d'=1$ (PM$_{2.5}$ predicted) \\
Latent dimension & $d_\ell=4$ \\
GRU hidden units & $64$ (1 layer) \\
GAT propagation steps & $K=2$ \\
Fourier transport coef. & $D>0$ (learned), $\mathbf{v}(t)$ (data)\\
ODE solver & Dormand–Prince \texttt{dopri5}\\
Solver tolerances & $\texttt{rtol}=\texttt{atol}=10^{-5}$ \\
MC–ODE trajectories & $S=\texttt{n\_traj\_samples}=3$ \\
Fusion gate & $\boldsymbol{\alpha}\in[0,1]^{N\times d_\ell}$\\
Physics loss weight & $\lambda_{\text{phys}}=10^{-2}$ \\
Evidential loss weight & $\lambda_{\text{unc}}=10^{-3}$ \\
Optimizer & Adam, lr $=5\times10^{-4}$ \\
LR schedule & Step decay (0.5 every 20 epochs)\\
Gradient clipping & $\|\nabla\|_2 \le 5$ \\
Graph operators & diffusion $\mathbf{L}(t)$, advection $\mathbf{M}_A(\mathbf{v}_t)$  \\
Filter type & \texttt{diff} / \texttt{adv} / \texttt{diff\_adv} \\
\bottomrule
\end{tabular}
\end{sc}
\end{center}
\vskip -0.1in
\label{tab:configuration}
\end{table*}

\section{Standing Assumptions (Theory)}
\label{sec:assumptions}

\begin{enumerate}
    \item \textbf{Well-posedness.} The vector field $f_\theta$ is globally Lipschitz with linear growth, ensuring unique ODE flows (\emph{Theorem~\ref{thm:wellposed}}).
    
    \item \textbf{Transport core.} The graph Laplacian $\mathbf{L}(t) \succeq 0$ is positive semidefinite and time-varying, and the advection operator $\mathbf{M}_A(\mathbf{v}_t)$ satisfies $\mathbf{M}_A(\mathbf{v}_t)^\top = -\mathbf{M}_A(\mathbf{v}_t)$ (skew-symmetric), yielding dissipative-unitary dynamics (\emph{Lemmas~\ref{lem:diffusion}, \ref{lem:advection}}).
    
    \item \textbf{Spectral prior.} Fourier multipliers satisfy $|H(\mathbf{k})| \leq 1$ for $D > 0$, ensuring nonexpansive spectral transport (\emph{Lemma~\ref{lem:fourier}}).
    
    \item \textbf{Residual size.} The data-driven residual $\Phi_\theta(\mathbf{z},\mathbf{E}_t)$ is $L_\Phi$-Lipschitz in $\mathbf{z}$; exponential stability on the mean-free subspace holds if $L_\Phi < D\lambda_{\min}^+(\mathbf{L}(t))$, where $\lambda_{\min}^+(\mathbf{L}(t))$ is the smallest positive eigenvalue of the time-varying Laplacian (\emph{Corollary~\ref{cor:stable}}).
    
    \item \textbf{Integrator.} Under $C^5$ smoothness of $f_\theta$, the Dormand-Prince method (dopri5) with $\texttt{rtol}=10^{-5}$ and $\texttt{atol}=10^{-5}$ achieves $O(h^4)$ global error (\emph{Proposition~\ref{prop:dopri5}}).
    
    \item \textbf{Fusion \& UQ.} The fusion MSE is upper bounded by convex combination of branch errors, and MC-ODE variance decomposition with $S=\texttt{n\_traj\_samples}$ trajectories obeys the law of total variance (\emph{Theorem~\ref{thm:mse}, Proposition~\ref{prop:ltv}}).
\end{enumerate}

Together, these assumptions guarantee that NeuroDDAF dynamics are well-posed, stable, and consistent with physical transport priors while maintaining the expressivity needed for complex air quality forecasting.

\section{Theoretical Results}
\label{app:theory}
For a dynamic graph $\mathcal{G}(t) = (\mathcal{V}, \mathcal{E}_t)$ with $N$ nodes (monitoring stations), let $\mathbf{L}(t) \in \mathbb{R}^{N \times N}$ denote the time-varying graph Laplacian, $\mathbf{M}_A(\mathbf{v}_t) \in \mathbb{R}^{N \times N}$ the wind-aware advection operator, and $\|\cdot\|$ the Euclidean norm. The spatial Fourier transform is denoted by $\mathcal{F}$; the graph Fourier transform corresponds to expansion in $\mathbf{L}(t)$'s eigenbasis. The encoder output is $\mathbf{E}_t \in \mathbb{R}^{N \times d_e}$, and the latent state is $\mathbf{z}(t) \in \mathbb{R}^{N \times d_\ell}$, where $d_\ell$ is the latent dimension. We write the NeuroDDAF vector field as:
\begin{equation}
\label{eq:app-field}
\begin{split}
f_\theta(\mathbf{z}, t, \mathbf{E}_t, \mathbf{v}(t), \mathcal{G}(t)) 
&= \underbrace{\mathbf{H}_{\mathrm{phys}}}_{\text{graph-based transport}} \\
&\quad + \underbrace{\Psi_{\mathrm{spec}}(\mathbf{z}, t)}_{\text{spectral transport}} 
+ \underbrace{\Phi_\theta(\mathbf{z}, \mathbf{E}_t)}_{\text{data-driven residual}}
\end{split}
\end{equation}
where $\mathbf{H}_{\mathrm{phys}} = \boldsymbol{\alpha} \odot \mathbf{H}_{\mathrm{diff}} + (1-\boldsymbol{\alpha}) \odot \mathbf{H}_{\mathrm{adv}}$ is the fused physics-based transport from equations (12)-(13) in the main text.

\subsection{Existence and Uniqueness of the Neural ODE Flow}
\label{app:wellposed}

\begin{assumption}[Lipschitz and linear growth]
\label{ass:lipschitz}
There exist constants $L_f, C_f > 0$ such that for all $t$ and all $\mathbf{z}, \mathbf{z}' \in \mathbb{R}^{N \times d_\ell}$:
\begin{equation}
\label{eq:lipschitz}
\begin{aligned}
\|f_\theta(\mathbf{z}, t, \mathbf{E}_t, \mathbf{v}(t), \mathcal{G}(t))
&- f_\theta(\mathbf{z}', t, \mathbf{E}_t, \mathbf{v}(t), \mathcal{G}(t))\| \\
&\le L_f \|\mathbf{z} - \mathbf{z}'\|
\end{aligned}
\end{equation}
and
\begin{equation}
\label{eq:linear-growth}
\|f_\theta(\mathbf{z}, t, \mathbf{E}_t, \mathbf{v}(t), \mathcal{G}(t))\|
\le C_f \big(1 + \|\mathbf{z}\|\big)
\end{equation}
This holds when $\|\mathbf{L}(t)\|, \|\mathbf{M}_A(\mathbf{v}_t)\|$ are bounded, activation functions are 1-Lipschitz (e.g., $\tanh$, ReLU), and network weights are bounded through normalization.
\end{assumption}

\begin{theorem}[Well-posedness]
\label{thm:wellposed}
Under Assumption~\ref{ass:lipschitz}, for any initial condition $\mathbf{z}(t_0) = \mathbf{z}_0$, the initial value problem $\frac{d\mathbf{z}(t)}{dt} = f_\theta(\mathbf{z}, t, \mathbf{E}_t, \mathbf{v}(t), \mathcal{G}(t))$ admits a unique, maximal, absolutely continuous solution on some interval containing $t_0$. If $f_\theta$ is globally Lipschitz, the solution exists and is unique for all $t \in \mathbb{R}$.
\end{theorem}

\begin{proof}
Follows from the Picard–Lindelöf (Cauchy–Lipschitz) theorem under Assumption~\ref{ass:lipschitz}. The linear growth condition ensures that solutions do not blow up in finite time.
\end{proof}

\subsection{Dissipation, Unitarity, and Stability of the Transport Priors}
\label{app:stability}

\begin{lemma}[Diffusion is contractive]
\label{lem:diffusion}
Let $\mathbf{L}(t) \succeq 0$ be a time-varying graph Laplacian. Then $e^{-tD\mathbf{L}(t)}$ is a contraction in $\ell_2$: for all $t \ge 0$, 
\begin{equation}
\|e^{-tD\mathbf{L}(t)}\| \le 1,
\end{equation}
and
\begin{equation}
\frac{d}{dt}\|\mathbf{z}(t)\|^2 = -2D\,\mathbf{z}(t)^\top \mathbf{L}(t)\mathbf{z}(t) \le 0
\end{equation}
\end{lemma}

\begin{proof}
Since $\mathbf{L}(t)$ is symmetric positive semidefinite for each $t$, it is diagonalizable with real eigenvalues $\lambda_i(t) \ge 0$. Hence $\|e^{-tD\mathbf{L}(t)}\| = \max_i e^{-tD\lambda_i(t)} \le 1$. The energy identity follows from the symmetry and positive semidefiniteness of $\mathbf{L}(t)$:
\begin{equation}
\frac{d}{dt}\|\mathbf{z}(t)\|^2 = 2\mathbf{z}(t)^\top \dot{\mathbf{z}}(t) = -2D\,\mathbf{z}(t)^\top \mathbf{L}(t)\mathbf{z}(t) \le 0
\end{equation}
\end{proof}

\begin{lemma}[Advection is norm-preserving]
\label{lem:advection}
If $\mathbf{M}_A(\mathbf{v}_t)$ is skew-symmetric ($\mathbf{M}_A(\mathbf{v}_t)^\top = -\mathbf{M}_A(\mathbf{v}_t)$), then $e^{t\mathbf{M}_A(\mathbf{v}_t)}$ is orthogonal and $\|e^{t\mathbf{M}_A(\mathbf{v}_t)}\| = 1$. Moreover, $\frac{d}{dt}\|\mathbf{z}(t)\|^2 = 0$ for $\dot{\mathbf{z}} = \mathbf{M}_A(\mathbf{v}_t)\mathbf{z}$.
\end{lemma}

\begin{proof}
For any skew-symmetric matrix $\mathbf{M}_A(\mathbf{v}_t)$, the matrix exponential $e^{t\mathbf{M}_A(\mathbf{v}_t)}$ is orthogonal. The norm preservation follows from:
\begin{equation}
\label{eq:advection-energy}
\begin{aligned}
\frac{d}{dt}\|\mathbf{z}(t)\|^2
&= 2\,\mathbf{z}(t)^\top \mathbf{M}_A(\mathbf{v}_t)\mathbf{z}(t) \\
&= \mathbf{z}(t)^\top \!\big(\mathbf{M}_A(\mathbf{v}_t)
     + \mathbf{M}_A(\mathbf{v}_t)^\top\big)\mathbf{z}(t) \\
&= 0
\end{aligned}
\end{equation}
\end{proof}

\begin{lemma}[Fourier prior is nonexpansive for $D > 0$]
\label{lem:fourier}
In the spectral domain, over a time step $\Delta t$ the Fourier multiplier is:
\begin{equation}
H(\mathbf{k}) = \exp\left(-D\|\mathbf{k}\|^2\Delta t\right) e^{-i(\mathbf{v}(t) \cdot \mathbf{k})\Delta t}
\end{equation}
Then $|H(\mathbf{k})| \le 1$ for all $\mathbf{k}$, with equality only if $D = 0$ or $\|\mathbf{k}\| = 0$.
\end{lemma}

\begin{proof}
Since $|e^{-i(\mathbf{v}(t) \cdot \mathbf{k})\Delta t}| = 1$ and $\exp\left(-D\|\mathbf{k}\|^2\Delta t\right) \le 1$ for $D, \Delta t \ge 0$, we have $|H(\mathbf{k})| \le 1$. Equality holds only when the diffusion term is identity, i.e., $D = 0$ or $\|\mathbf{k}\| = 0$.
\end{proof}

\begin{theorem}[Dissipativity of the linear transport core]
\label{thm:dissipative}
Consider $\dot{\mathbf{z}} = -(D\mathbf{L}(t) - \mathbf{M}_A(\mathbf{v}_t))\mathbf{z}$ with $\mathbf{L}(t) \succeq 0$ and $\mathbf{M}_A(\mathbf{v}_t)^\top = -\mathbf{M}_A(\mathbf{v}_t)$. Then:
\begin{equation}
\frac{d}{dt}\|\mathbf{z}(t)\|^2 = -2D\,\mathbf{z}^\top\mathbf{L}(t)\mathbf{z} \le 0
\end{equation}
and the flow is nonexpansive in $\ell_2$. If $\lambda_{\min}^+(\mathbf{L}(t)) > 0$ and $D > 0$, the origin is exponentially stable on the orthogonal complement of the constant eigenvector.
\end{theorem}

\begin{proof}
Combining Lemmas~\ref{lem:diffusion} and \ref{lem:advection}:
\begin{equation}
\label{eq:dissipation}
\begin{aligned}
\frac{d}{dt}\|\mathbf{z}(t)\|^2
&= -2D\,\mathbf{z}^\top \mathbf{L}(t)\mathbf{z}
   + \mathbf{z}^\top \!\big(\mathbf{M}_A(\mathbf{v}_t)
   + \mathbf{M}_A(\mathbf{v}_t)^\top\big)\mathbf{z} \\
&= -2D\,\mathbf{z}^\top \mathbf{L}(t)\mathbf{z} \\
&\le 0
\end{aligned}
\end{equation}
When $\lambda_{\min}^+(\mathbf{L}(t)) > 0$ and $D > 0$, the system exhibits exponential stability on the subspace orthogonal to the kernel of $\mathbf{L}(t)$
\end{proof}

\begin{corollary}[Stability with residuals]
\label{cor:stable}
Let $\Phi_\theta(\mathbf{z},\mathbf{E}_t)$ be $L_\Phi$-Lipschitz in $\mathbf{z}$. Then:
\begin{equation}
\frac{d}{dt}\|\mathbf{z}\|^2 \le -2D\lambda_{\min}^+(\mathbf{L}(t))\|\mathbf{z}_\perp\|^2 + 2L_\Phi\|\mathbf{z}\|^2
\end{equation}
where $\mathbf{z}_\perp$ is the projection onto the mean-free subspace. If $L_\Phi < D\lambda_{\min}^+(\mathbf{L}(t))$, the system is exponentially stable on the mean-free subspace.
\end{corollary}

\begin{proof}
Decompose $\mathbf{z} = \mathbf{z}_\parallel + \mathbf{z}_\perp$, where $\mathbf{z}_\parallel$ is in the kernel of $\mathbf{L}(t)$ (constant vectors) and $\mathbf{z}_\perp$ is mean-free. Since $\mathbf{L}(t)\mathbf{z}_\parallel = 0$, we have:
\begin{equation}
\frac{d}{dt}\|\mathbf{z}_\perp\|^2 \le -2D\lambda_{\min}^+(\mathbf{L}(t))\|\mathbf{z}_\perp\|^2 + 2L_\Phi\|\mathbf{z}\|^2
\end{equation}
Under the condition $L_\Phi < D\lambda_{\min}^+(\mathbf{L}(t))$, Gronwall's inequality implies exponential decay of $\|\mathbf{z}_\perp\|^2$ when $\|\mathbf{z}_\parallel\|$ is bounded.
\end{proof}

\subsection{Consistency of the Physics Residual Loss}
\label{app:physloss}

Let $\mathcal{P}$ denote the continuous advection-diffusion operator $\frac{\partial c}{\partial t} = D\nabla^2 c - \mathbf{v} \cdot \nabla c$, and $\mathcal{P}_h$ its Fourier-domain discretization used in training. Define the physics residual loss:
\begin{equation}
\mathcal{L}_{\mathrm{PDE}} = \|\mathcal{P}_h[\hat{\mathbf{y}}]\|_2^2
\end{equation}

\begin{proposition}[Residual consistency]
\label{prop:consistency}
Suppose $\mathcal{P}_h$ is a consistent discretization of $\mathcal{P}$ of order $q$ and $\hat{\mathbf{y}}_h \to y$ in the appropriate norm as $h \to 0$. Then $\mathcal{L}_{\mathrm{PDE}}(\hat{\mathbf{y}}_h) \to 0$ if and only if $\mathcal{P}[y] = 0$.
\end{proposition}

\begin{proof}
By consistency, $\|\mathcal{P}_h[\hat{\mathbf{y}}_h] - \mathcal{P}[y]\| = O(h^q)$. If $\mathcal{P}[y] = 0$, the residual vanishes in the limit. Conversely, if the residual $\to 0$, then $\mathcal{P}[y] = 0$ by consistency of the discretization.
\end{proof}

\subsection{Fusion Bounds and Diversity Term}
\label{app:fusion}

Let $y$ be the target value, $a = \hat{y}_{\mathrm{spec}}$ the physics-based (spectral) prediction, $b = \hat{y}_{\mathrm{nn}}$ the neural ODE prediction, and $\hat{y}_\gamma = \gamma a + (1 - \gamma)b$ the fused prediction with $\gamma \in [0,1]$ (applied pointwise), consistent with the evidential fusion gate.

\begin{lemma}[MAE bound (convexity)]
\label{lem:mae}
\begin{equation}
|y - \hat{y}_\gamma| \le \gamma|y - a| + (1 - \gamma)|y - b|
\end{equation}
\end{lemma}

\begin{proof}
By the triangle inequality:
\begin{equation}
\label{eq:mae-bound}
\begin{aligned}
|y - \gamma a - (1 - \gamma)b|
&= |\gamma(y - a) + (1 - \gamma)(y - b)| \\
&\le \gamma|y - a| + (1 - \gamma)|y - b|
\end{aligned}
\end{equation}
\end{proof}

\begin{theorem}[MSE decomposition and bound]
\label{thm:mse}
Let $e_a = y - a$, $e_b = y - b$. Then:
\begin{align}
(y - \hat{y}_\gamma)^2 &= \gamma e_a^2 + (1 - \gamma)e_b^2 - \gamma(1 - \gamma)(a - b)^2 \label{eq:mse-equality} \\
&\le \gamma e_a^2 + (1 - \gamma)e_b^2 \label{eq:mse-ineq}
\end{align}
Thus, fusion never exceeds the convex combination of branch MSEs; larger branch \emph{diversity} $(a - b)^2$ strictly lowers the fused error.
\end{theorem}

\begin{proof}
Expand the squared error:
\begin{equation}
\label{eq:mse-expand}
\begin{aligned}
(y - \hat{y}_\gamma)^2
&= (\gamma e_a + (1 - \gamma)e_b)^2 \\
&= \gamma^2 e_a^2 + (1 - \gamma)^2 e_b^2
   + 2\gamma(1 - \gamma)e_a e_b
\end{aligned}
\end{equation}
Using the identity $e_a e_b = \frac{1}{2}(e_a^2 + e_b^2 - (a - b)^2)$ and rearranging terms yields equation~\eqref{eq:mse-equality}. The inequality~\eqref{eq:mse-ineq} follows since $-\gamma(1 - \gamma)(a - b)^2 \le 0$.
\end{proof}

\begin{corollary}[Oracle gate]
\label{cor:oracle}
The oracle gate $\gamma^\star = \mathbb{1}[e_a^2 < e_b^2]$ yields $(y - \hat{y}_{\gamma^\star})^2 = \min\{e_a^2, e_b^2\}$. A learned gate minimizing an upper bound in \eqref{eq:mse-ineq} recovers the oracle in the limit of perfect estimation.
\end{corollary}

\subsection{Uncertainty Decomposition and Evidential Regression}
\label{app:uncertainty}

Let $\hat{\mathbf{Y}}^{(s)}$ be $S$ fused trajectories from the Monte Carlo ODE ensemble with $\texttt{n\_traj\_samples}=S$, and $\bar{\mathbf{Y}} = \frac{1}{S}\sum_{s=1}^S \hat{\mathbf{Y}}^{(s)}$ the ensemble mean.

\begin{proposition}[Law of total variance]
\label{prop:ltv}
For each coordinate, the total variance decomposes as:
\begin{equation}
\mathrm{Var}(Y) = \mathbb{E}[\mathrm{Var}(Y \mid S)] + \mathrm{Var}(\mathbb{E}[Y \mid S])
\end{equation}
Estimating with sample moments gives:
\begin{equation}
\mathbf{V}_{\mathrm{tot}} = \underbrace{\frac{1}{S}\sum_{s=1}^S \boldsymbol{\Sigma}^{(s)}_{\mathrm{ale}}}_{\text{aleatoric uncertainty}} + \underbrace{\frac{1}{S}\sum_{s=1}^S (\hat{\mathbf{Y}}^{(s)} - \bar{\mathbf{Y}})^2}_{\text{epistemic uncertainty}}
\end{equation}
where $\boldsymbol{\Sigma}^{(s)}_{\mathrm{ale}}$ is the heteroscedastic (aleatoric) variance produced by the evidential head when enabled.
\end{proposition}

\textbf{Evidential (Normal-Inverse-Gamma) head for scalar outputs.}
Let $y \in \mathbb{R}$ be a scalar target and $(\mu, \lambda, \alpha, \beta)$ the predictive parameters from the evidential head with constraints $\alpha > 1$, $\beta > 0$, $\lambda > 0$. The predictive distribution follows a Student-$t$ distribution with mean $\mu$ and variance $\beta/(\lambda(\alpha - 1))$.

\subsection{Convergence and Complexity of the ODE Integrator}
\label{app:ode-conv}

\begin{assumption}[Local smoothness]
\label{ass:smooth}
The vector field $f_\theta(\cdot, t, \mathbf{E}_t, \mathbf{v}(t), \mathcal{G}(t))$ is $C^5$ in a neighborhood of the trajectory with bounded derivatives up to order 5.
\end{assumption}

\begin{proposition}[dopri5 error order]
\label{prop:dopri5}
Under Assumption~\ref{ass:smooth}, the Dormand-Prince (5,4) method with adaptive stepsizes and tolerances $\texttt{rtol}=10^{-5}$, $\texttt{atol}=10^{-5}$ achieves global error $O(h^4)$, where $h$ is the typical stepsize determined by the tolerances.
\end{proposition}

\begin{proposition}[Per-step computational cost]
\label{prop:cost}
Each evaluation of $f_\theta$ costs $O(K|\mathcal{E}_t|d_\ell + N d_\ell \log N)$, where $K$ is the number of graph propagation steps (default $K=2$), $|\mathcal{E}_t|$ is the number of edges in the dynamic graph (for GAT/Laplacian operations), and $N d_\ell \log N$ comes from FFT/IFFT operations per latent channel.
\end{proposition}

\subsection{Coverage of Prediction Intervals (Gaussian/Student-$t$)}
\label{app:coverage}

\begin{proposition}[Asymptotic nominal coverage]
\label{prop:coverage}
If the per-coordinate predictive distribution is Gaussian $\mathcal{N}(\bar{y}, \sigma^2)$ (or Student-$t$ with $\nu$ degrees of freedom), the $(1 - \alpha)$ prediction interval $\bar{y} \pm z_{1-\alpha/2}\sigma$ (or $\bar{y} \pm t_{1-\alpha/2,\nu}\sigma$) has nominal coverage under correct specification. Using $\sigma^2 = \mathbf{V}_{\mathrm{tot}}$ with MC-ODE ensembles yields asymptotically nominal coverage as $S \to \infty$ under exchangeability and finite moment assumptions.
\end{proposition}

\subsection{Putting It Together: A Stability–Expressivity Tradeoff}
\label{app:tradeoff}

\begin{theorem}[Overall stability condition]
\label{thm:overall}
Let $\mathbf{L}(t) \succeq 0$ with $\lambda_{\min}^+(\mathbf{L}(t)) > 0$, $\mathbf{M}_A(\mathbf{v}_t)^\top = -\mathbf{M}_A(\mathbf{v}_t)$, and $\Phi_\theta(\mathbf{z},\mathbf{E}_t)$ be $L_\Phi$-Lipschitz. Then the linear transport core is dissipative (Theorem~\ref{thm:dissipative}); the full system is exponentially stable on the mean-free subspace if $L_\Phi < D\lambda_{\min}^+(\mathbf{L}(t))$. Increasing $D$ or graph connectivity (larger $\lambda_{\min}^+$) improves stability but may oversmooth; the data-driven residual $\Phi_\theta$ restores expressivity while maintaining stability under the Lipschitz condition.
\end{theorem}

\textbf{Remark (Discrete-time view).}
With explicit Euler discretization, a sufficient stability condition for the spectral step is $\Delta t \le 2/(D\|\mathbf{k}\|^2_{\max})$. The adaptive dopri5 method used in NeuroDDAF automatically handles stability through adaptive step sizing while preserving the dissipativity guarantees of the continuous system.

\section{Case Study Extensions}

\subsection{Geospatial Representation of PM$_{2.5}$ Transport}

\begin{figure*}[t]
    \centering
    \includegraphics[width=0.80\linewidth]{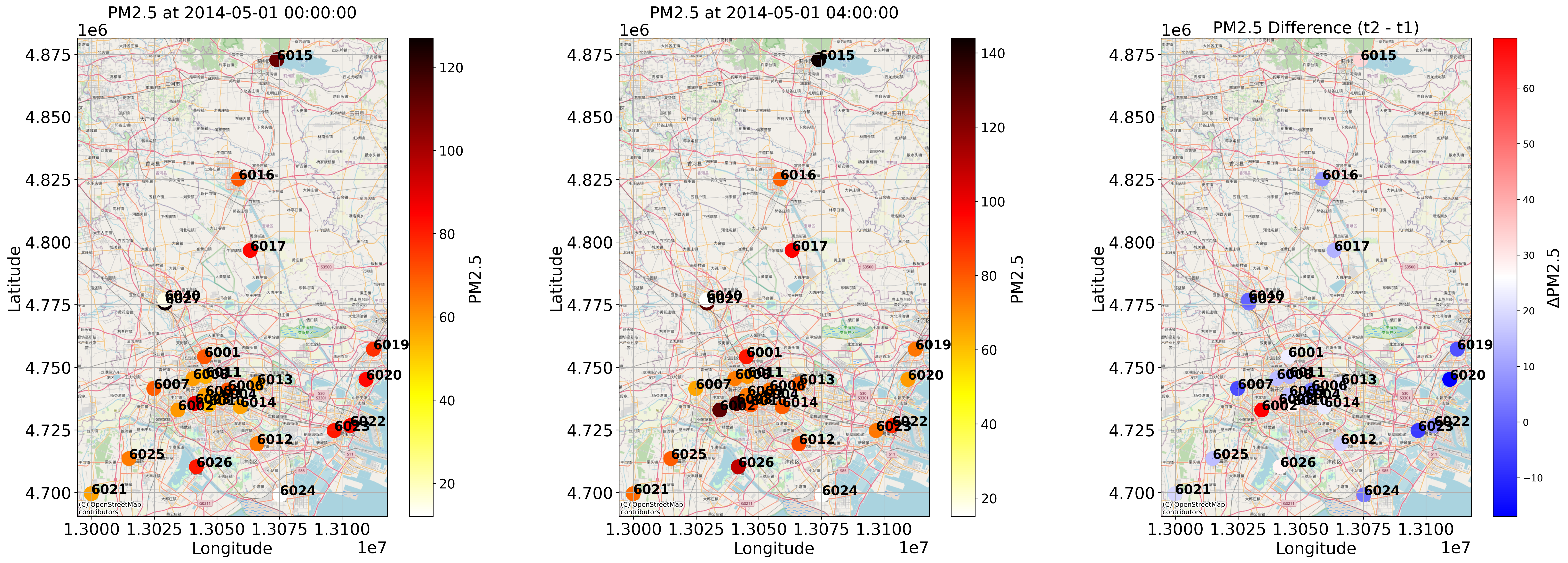}
    \caption{Geospatial representation of PM$_{2.5}$ in Tianjin at 00:00 (left) and 04:00 (middle) on May 1, 2014, and their difference (right). Station identifiers, wind vectors, and an OpenStreetMap basemap provide geographic context to the transport patterns.}
    \label{fig:pm25_appendix}
\end{figure*}

To complement the case study presented in Section~IV.E, we provide an extended visualization incorporating explicit geographic coordinates and a cartographic basemap. Figure~\ref{fig:pm25_appendix} shows PM$_{2.5}$ concentrations in Tianjin at 00:00 and 04:00 on May 1, 2014, together with their difference field. In contrast to the simplified station-value maps in the main text, this version overlays the measurements directly on a city map, with latitude and longitude as axes and OpenStreetMap tiles as a baselayer.

Each monitoring station is labeled with its identifier, allowing for unambiguous geographic reference. Wind vectors are plotted alongside pollutant values, providing a more intuitive understanding of transport relative to local landmarks and urban structure. This spatially explicit representation confirms the findings of the main analysis: downwind stations record increases in PM$_{2.5}$, while upwind stations show decreases, consistent with wind-driven redistribution of pollutants.

\subsection{Transport Regime Classification}
\label{sec:regime}
\begin{figure*}[b]
    \centering
    \includegraphics[width=0.9\linewidth]{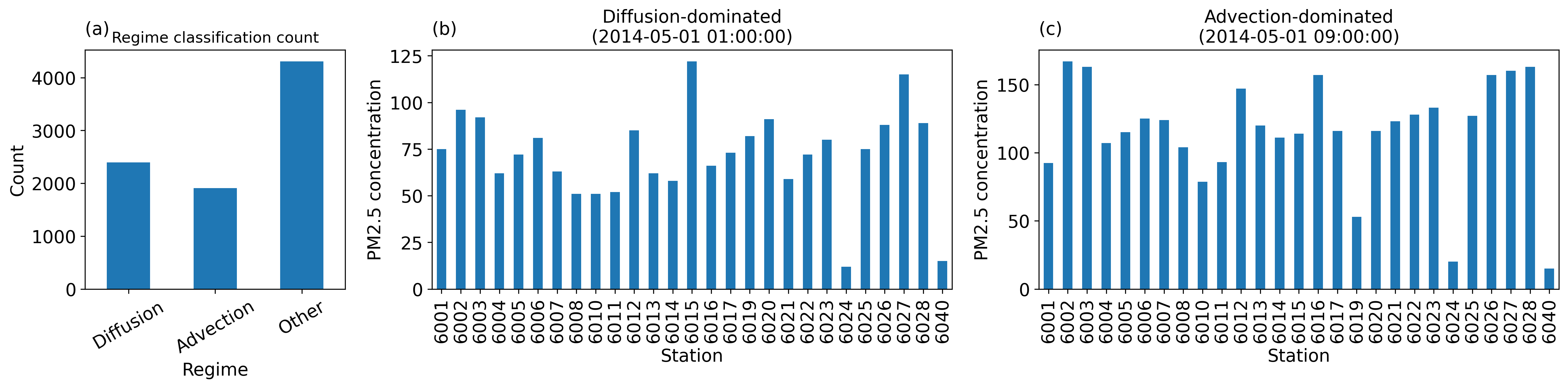}
    \caption{PM$_{2.5}$ regime classification. (a) Counts of diffusion-, advection-, and other regimes, where ``other'' corresponds to time steps with intermediate wind speeds or weak gradients that do not meet thresholds for either transport regime. (b) Example of a diffusion-dominated case with smoother gradients. (c) Example of an advection-dominated case with directional variability under strong winds.}
    \label{fig:regime_panels}
\end{figure*}

To quantify the role of wind and diffusion more systematically, we classified each time step into three transport regimes: diffusion-dominated, advection-dominated, and other. The classification was based on thresholds of average wind speed and spatial PM$_{2.5}$ gradients (Figure~\ref{fig:regime_panels}a). Among all time steps, the majority fell into the ``other'' category, while advection- and diffusion-dominated episodes together accounted for a substantial fraction of cases.

Panels (b) and (c) illustrate representative diffusion- and advection-dominated regimes. In the diffusion-dominated case, PM$_{2.5}$ concentrations appear relatively homogeneous, consistent with slow redistribution under weak winds. In the advection-dominated case, stronger wind transport introduces sharper directional variability, with certain downwind stations registering elevated concentrations compared to less-affected regions. While the overall spatial patterns in panels (b) and (c) remain broadly similar due to the relatively dense monitoring network, the contrast between smoother gradients in diffusion-dominated episodes and sharper directional transport in advection-dominated episodes becomes clearer in synthetic experiments. 

These results emphasize that wind plays an important role in shaping spatio-temporal PM$_{2.5}$ dynamics. Diffusion tends to stabilize pollutant fields and smooth gradients, whereas advection enhances variability and can transport plumes along prevailing wind directions. This case study highlights both the value and the challenge of explicitly modeling wind-driven dynamics in forecasting frameworks such as NeuroDDAF.

